\documentclass[runningheads]{llncs}

\usepackage{eccv}

\usepackage{eccvabbrv}

\usepackage{graphicx}
\usepackage{booktabs}

\usepackage[accsupp]{axessibility}  

\usepackage{hyperref}

\definecolor{cvprblue}{rgb}{0.21,0.49,0.74}

\usepackage{graphicx}
\usepackage{xspace}
\usepackage{booktabs}
\usepackage{wrapfig}
\usepackage{array}
\usepackage{tikz}
\usepackage{pgfplots}
\usepackage{colortbl}
\usepackage[accsupp]{axessibility}  %
\usepackage{orcidlink}
\usepackage{tabularx}
\usepackage{multirow}
\usepackage{tcolorbox}
\usepackage{tikz}
\usepackage{pifont}
\usepackage{microtype}
\usepackage{fontawesome}
\usepackage[toc,page]{appendix}
\usepackage{bm}
\usepackage{gradient-text}
\usepackage{textcomp}
\usepackage{gensymb}
\usepackage{lipsum}
\usepackage{enumitem}
\usepackage{duckuments}
\usepackage{xcolor}
\usepackage{caption}
\usepackage[capitalize]{cleveref}
\crefname{section}{Sec.}{Secs.}
\Crefname{section}{Section}{Sections}
\Crefname{table}{Table}{Tables}
\crefname{table}{Tab.}{Tabs.}

\usepackage{caption}
\newcommand{\pheading}[1]{\medskip\noindent\textbf{#1}}

\definecolor{bestgreen}{RGB}{153,200,76}
\definecolor{worstred}{RGB}{192,0,0}

\definecolor{cbad}{HTML}{FFD0D0}
\definecolor{cmedium}{HTML}{FFF0D0}
\definecolor{cgood}{HTML}{90C060}

\usepackage{pgfplots}
\pgfplotsset{compat=1.18}

\newlength{\savewidth}
\newcommand{\shline}{\noalign{\global\savewidth\arrayrulewidth \global\arrayrulewidth 1pt}\hline\noalign{\global\arrayrulewidth\savewidth}}

\definecolor{DeltaColor}{rgb}{0.039,0.73,0.71}
\definecolor{SigmaColor}{rgb}{0.98,0.45,0.0}
\definecolor{AlphaColor}{rgb}{0,0,0.8}
\definecolor{BetaColor}{rgb}{0.8,0,0.8}
\definecolor{GammaColor}{rgb}{0.514,0.34,0.224}
\definecolor{EpsilonColor}{rgb}{0.353,0.725,0.906}
\definecolor{PurpleColor}{HTML}{9839ff}
\definecolor{RedColor}{rgb}{0.949,0.275, 0.224}
\definecolor{citecolor}{HTML}{0071bc}

\definecolor{deepred}{HTML}{940000}

\newcommand{\modelname}{\mbox{OmniFit}\xspace}

\newcommand{\longname}{Multi-modal 3D Body Fitting via Scale-agnostic Dense Landmark Prediction}

\definecolor{OrangeColor}{rgb}{0.914,0.541,0.0.141}
\definecolor{GreenColor}{rgb}{0.137,0.573,0.565}

\definecolor{HeadGreen}{RGB}{124,231,124}
\definecolor{BodyRed}{RGB}{255,124,124}
\definecolor{HandYellow}{RGB}{251,223,26}
\definecolor{Fitted}{RGB}{116,218,225}
\definecolor{OuterPoints}{RGB}{241,94,92}

\newcommand{\sota}{\mbox{state-of-the-art}\xspace}

\newcommand{\real}{\mathbb{R}}

\newcommand{\smpl}{\mbox{SMPL}\xspace}
\newcommand{\smplh}{\mbox{SMPL-H}\xspace}

\newcommand{\smplx}{\mbox{SMPL-X}\xspace}

\newcommand{\cape}{\mbox{CAPE}\xspace}
\newcommand{\ddress}{\mbox{4D-DRESS}\xspace}

\newcommand{\nicp}{\mbox{NICP}\xspace}
\newcommand{\arteq}{\mbox{ArtEq}\xspace}
\newcommand{\ipnet}{\mbox{IPNet}\xspace}
\newcommand{\ptf}{\mbox{PTF}\xspace}
\newcommand{\etch}{\mbox{ETCH}\xspace}

\newcommand{\vsmplx}{\mathbf{V}}

\newcommand{\shapecoeff}{\boldsymbol{\beta}}
\newcommand{\expressioncoeff}{\boldsymbol{\psi}}
\newcommand{\shapedim}{{\left| \shapecoeff \right|}}

\newcommand{\posecoeff}{\boldsymbol{\theta}}

\newcommand{\translcoeff}{\boldsymbol{t}}

\newcommand{\template}{\mathbf{\bar{T}}}
\newcommand{\restpose}{\mathbf{T}}

\newcommand{\joints}{J}

\newcommand{\blendweights}{\mathcal{W}}

\newcommand{\pcds}{\mathbf{X}}
\newcommand{\pcd}{\mathbf{x}}
\newcommand{\normpcds}{\tilde{\mathbf{X}}}
\newcommand{\lmks}{\mathbf{L}}
\newcommand{\lmk}{\mathbf{l}}
\newcommand{\lmkembeds}{\mathbf{Q}}
\newcommand{\pcdembeds}{\mathbf{E}}
\newcommand{\pcdembed}{\mathbf{e}}
\newcommand{\pcdfeats}{\mathbf{F_{P}}}
\newcommand{\lmkfeats}{\mathbf{F_{L}}}
\newcommand{\numlayers}{l}
\newcommand{\numgroup}{g}

\newcommand{\numpcds}{N}
\newcommand{\numlmks}{M}

\newcommand{\image}{\mathbf{I}}
\newcommand{\imagefeats}{\mathbf{F_{I}}}

\newcommand{\scalefactor}{\mathbf{S}}
\newcommand{\gtscalefactor}{\hat{\mathbf{S}}}
\newcommand{\scaletoken}{\mathbf{E[s]}}

\newcommand{\normpcdembed}{\tilde{\mathbf{e}}}

\newcommand{\gtlmks}{\hat{\mathbf{L}}}
\newcommand{\gtlmk}{\hat{\mathbf{l}}}
\newcommand{\optimlmk}{\tilde{\mathbf{l}}}

\usepackage{orcidlink}

\begin{document}

\title{\modelname: \longname} 
\author{Zeyu Cai$^{1}$, Yuliang Xiu$^{2}$, Renke Wang$^{4}$, Zhijing Shao$^{5}$, Xiaoben Li$^{2}$, Siyuan~Yu$^{2}$, Chao Xu$^{3}$, Yang Liu$^{3}$, Baigui Sun$^{3}$, Jian Yang$^{1}$, Zhenyu Zhang$^{1\dagger}$}

\institute{
    $^{1}$Nanjing University \quad $^{2}$Westlake University \quad $^{3}$iROOTECH \\
    $^{4}$Nanjing University of Science and Technology \\
    $^{5}$The Hong Kong University of Science and Technology (Guangzhou) \\
    $^{\dagger}$Corresponding Author \quad Project: \url{https://zcai0612.github.io/OmniFit}
}

\authorrunning{\modelname arXiv Preprint}
\titlerunning{\modelname arXiv Preprint}

\maketitle
\begin{minipage}{\textwidth}
    \centering
    \includegraphics[trim=000mm 000mm 000mm 000mm, clip=true, width=\linewidth]{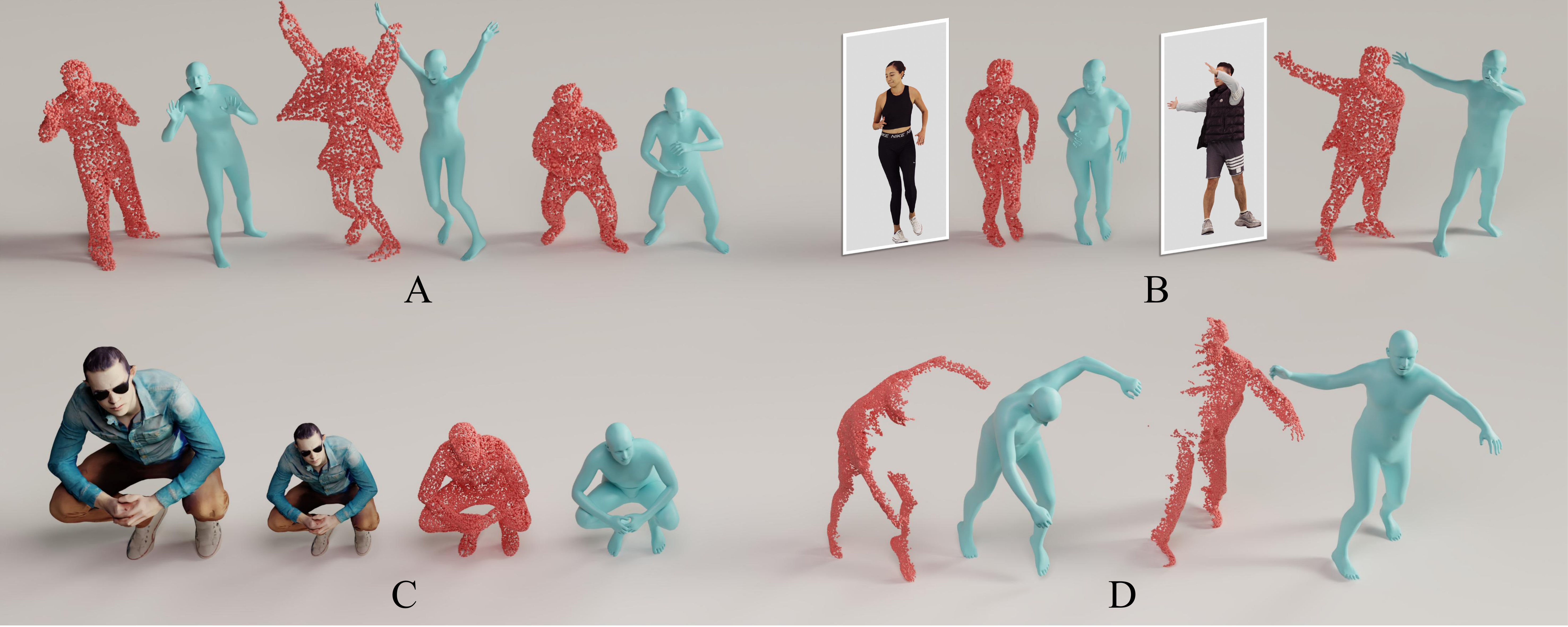}
    \vspace{-1.5em}
    \captionsetup{type=figure}
    \captionof{figure}{\scriptsize \textbf{\modelname handles diverse 3D human data sources and input modalities.}
    \textbf{(A)}~Raw point clouds from real-world captures.
    \textbf{(B)}~Point clouds jointly conditioned on a front-view RGB image.
    \textbf{(C)}~Scale-distorted AI-generated 3D human assets.    \textbf{(D)}~Partial point clouds reconstructed from depth maps.
    For each case, we display the \textcolor{OuterPoints}{input point cloud} alongside the \textcolor{Fitted}{fitted \smplx body}, demonstrating the robust fitting capability of \modelname across all modalities and data sources.
    \label{fig:teaser}}
\end{minipage}
\begin{abstract}
Fitting an underlying body model to 3D clothed human assets has been extensively studied, yet most approaches focus on either single-modal inputs such as point clouds or multi-view images alone, often requiring known metric scale—a constraint that is frequently unavailable, particularly for AIGC-generated assets where scale distortion is prevalent. We propose \modelname, where ``Omni'' signifies our method's ability to seamlessly handle diverse multi-modal inputs (\eg, full scans, partial depth, image captures) while remaining scale-agnostic across both real and synthetic assets. Our key innovation is a simple yet effective conditional transformer decoder that directly transforms surface points into dense body landmarks, which are subsequently used for \smplx parameter fitting. Additionally, an optional plug-and-play image adapter enriches geometric details with visual cues to address potential incompleteness. We further introduce a dedicated scale predictor to resize subjects into canonical proportions. Remarkably, \modelname substantially outperforms state-of-the-art methods by \textcolor{GreenColor}{57.1\%--80.9\%} across daily and loose clothing scenarios, making it the first body fitting method to surpass multi-view optimization baselines and the first to achieve \textbf{millimeter-level} accuracy on CAPE and \ddress benchmarks.

\end{abstract}

\section{Introduction}
\label{sec:intro}

Fitting an underlying parametric body model~\cite{SMPLX:2019} to a 3D clothed human asset is a fundamental task in computer vision and graphics, with broad applications in human animation~\cite{cao2024dreamavatar,qiu2025lhm,shao2025degas}, avatar creation~\cite{li2024pshuman,he2025magicman,qiu2025anigs,cai2024dreammapping,wang2025headevolver}, and AR/VR~\cite{chen2025taoavatar, cai2024interact360,wang2025magicscroll}. Previous body fitting methods have primarily focused on a single input modality, such as RGB images~\cite{easymocap,wang2025diffproxy}, point clouds~\cite{bhatnagar2020ipnet,wang2021ptf}, or depth maps~\cite{lascheit2025segfit}. However, in many real-world scenarios, complementary modalities are simultaneously available---for instance, RGBD cameras that jointly capture color and geometry~\cite{tao2021thuman2.1,wang20244ddress,ho2023customhumans}, or AI-based content creation tools that generate textured 3D human meshes~\cite{zhao2025hunyuan3d,xiang2025trellis,chen2025synchuman,cai2025up2you,wang2025headevolver}. 
These scenarios inspire us to leverage multi-modal sources together for more accurate and robust body fitting.

To this end, we propose \modelname, a unified framework for fitting \smplx body models to 3D clothed human assets across diverse input modalities and data sources. As illustrated in~\cref{fig:teaser}, the name ``\modelname'' reflects two key properties: (1)~it accepts \textit{multi-modal} inputs, seamlessly handling point clouds alone (A) or jointly with an optional RGB image (B), while gracefully accommodating scale-distorted (C) or partial (D) geometry; (2)~it generalizes to \textit{various} clothed human---whether from real-world captures, synthetic datasets, or AI-generated avatars---regardless of pose, body shape, clothing style, or scale.

Unlike traditional optimization-based fitting pipelines that require multi-view rendered images and cascade through multiple error-prone stages (\cref{sec:related}), \modelname is a 3D-native, learning-based approach built around three key components. At its core is a \textit{landmark predictor} (\cref{sec:method:lmk}), a conditional transformer decoder that directly estimates a set of predefined dense \smplx landmarks from an input point cloud. Inspired by~\cite{cuevas2025mamma}, this end-to-end design provides explicit and direct fitting targets for subsequent \smplx optimization, making it better suited for data-driven training than previous dense correspondence methods~\cite{bhatnagar2020ipnet,wang2021ptf}. Moreover, the freely configurable landmark distribution offers comprehensive coverage of fine-grained body regions such as the face and hands, overcoming the limitations of aggregation-based markers~\cite{li2025etch} that are uniformly distributed and struggle with missing or partial inputs.

To further leverage appearance information available in many 3D human assets, we introduce a \textit{plug-and-play image adapter} (\cref{sec:method:adapter}) that incorporates a front-view RGB image as an optional additional condition, improving landmark prediction without altering the base architecture. To handle the scale ambiguity that commonly arises in AI-generated assets, we additionally design a \textit{scale predictor} (\cref{sec:method:scale_pred}) that estimates a scale factor from a normalized point cloud and restores it to canonical human size prior to landmark prediction.

Our main contributions are:
\begin{itemize}
    \item We present \modelname, a unified multi-modal framework for \smplx body fitting that seamlessly handles point clouds, optional RGB images, and scale-distorted or partial inputs, generalizing to \textit{various} clothed human across real-world captures, synthetic datasets, and AI-generated assets.
    \item We introduce three novel modules---a conditional transformer decoder for dense \smplx landmark prediction, a plug-and-play image adapter for optional RGB fusion, and a scale predictor for resolving scale ambiguity in AI-generated assets---that together make up \modelname, enabling accurate, robust, and versatile body fitting.
    \item To our knowledge, \modelname is the first 3D body fitting method to achieve \textbf{millimeter-level} accuracy, reducing V2V and MPJPE errors by up to \textcolor{GreenColor}{75.5\%} and \textcolor{GreenColor}{80.9\%} over the second-best method \etch~\cite{li2025etch} on standard benchmarks, while surpassing leading multi-view optimization-based pipelines~\cite{easymocap,wang2025diffproxy}.
\end{itemize}

\section{Related Work}
\label{sec:related}


\pheading{Optimization-based Body Fitting.}
Early optimization-based methods for body fitting typically leverage the Iterative Closest Point (ICP) algorithm~\cite{chen1992objectregistration} or its variants~\cite{allen2003humanspace, pons2015dyna, zuffi2015stitched} to align a body template with the target geometry.
Modern optimization-based approaches~\cite{zhang2017buff,ma2020cape,patel2021agora,zheng2019deephuman,tao2021thuman2.1,rbh_reg,easymocap,wang2025diffproxy} instead operate on rendered multi-view RGB images through a multi-stage pipeline: the input scan is first rendered from multiple viewpoints, followed by 2D keypoint detection (\eg, via OpenPose~\cite{cao2019openpose} or AlphaPose~\cite{fang2022alphapose}); these 2D detections are then lifted to 3D keypoints via multi-view triangulation; finally, \smplx parameters are optimized to align with the reconstructed 3D keypoints and point cloud, with optional constraints such as skin segmentation~\cite{antic2024close,Gong2019Graphonomy} or self-intersection penalties~\cite{mueller2021tuch}.
However, errors in each stage propagate through the pipeline and compound into significant inaccuracies in the final result.
Furthermore, these methods fundamentally depend on multi-view rendering, which is inapplicable to untextured meshes or partial point clouds.

\pheading{Learning-based Body Fitting.}
Learning-based methods offer efficient alternatives by leveraging large-scale 3D human datasets~\cite{mahmood2019amass, wang20244ddress, ma2020cape, bertiche2020cloth3d} and deep neural networks~\cite{qi2017pointnet,qi2017pointnet++,zaheer2017deepsets,thomas2019kpconv,zhao2021pointtransformer,wu2022ptv2,yu2022pointbert,wu2024ptv3} tailored for point cloud processing. These approaches either provide initialization for subsequent fitting~\cite{wang2021ptf, bhatnagar2020ipnet,bhatnagar2020loopreg} or directly regress statistical body model parameters~\cite{feng2023arteq,jiang2019skeleton,liu2021votehmr}. Since direct parameter regression is challenging, most methods rely on intermediate proxy representations, such as joint features~\cite{jiang2019skeleton,liu2021votehmr}, correspondence maps~\cite{bhatnagar2020loopreg,wang2021ptf}, part segmentation~\cite{wang2021ptf,feng2023arteq}, or sparse markers~\cite{li2025etch}. \modelname follows the learning-based paradigm and uses predefined dense landmarks as the fitting proxy. Unlike prior methods that take only point clouds as input, \modelname optionally incorporates a front-view RGB rendering as an additional condition, enabling it to exploit the full information available in 3D human assets.

\pheading{Correspondence Prediction for Body Fitting.}
Correspondence prediction is a key step in human body fitting. Explicit dense correspondence methods~\cite{bhatnagar2020ipnet,bhatnagar2020loopreg,wang2021ptf,marin24nicp} query pointwise correspondence features from a learned implicit field, establishing a mapping from input points to their corresponding semantic regions on the body template. As dense sampling is computationally expensive, \etch~\cite{li2025etch} instead aggregates dense correspondences into a compact set of markers placed at fixed locations on the body template, each representing a cluster center of correspondence features. This design simplifies training but struggles with partial inputs, as missing regions may cause markers to be lost. \modelname similarly maps input point clouds to predefined landmarks; however, unlike \etch, landmark locations are not aggregated from dense correspondences. Instead, \modelname conditions on the input points to directly decode landmark coordinates, with part correspondence implicitly learned and encoded in cross-attention maps. 
\section{Method}
\label{sec:method}

Given a point cloud $\pcds = \{\pcd_{i} \in \real^{3}\}_{i=1}^{\numpcds}$ of a 3D clothed human, which may be randomly sampled from a mesh, derived from 3DGS point positions, or directly captured by RGBD sensors, our goal is to reconstruct the underlying human body in \smplx format (\cref{sec:method:preli}). To achieve this, our core solution is fitting \smplx parameters to a set of predefined dense landmarks $\lmks = \{\lmk_{i} \in \real^{3}\}_{i=1}^{\numlmks}$ estimated from the input point cloud $\pcds$ (\cref{sec:method:lmk}). To further use the geometric information encoded in the 3D human, we introduce a plug-and-play adapter that enhances landmark prediction with an additional image condition (\cref{sec:method:adapter}). In addition, to address scale distortion in point cloud---particularly those derived from generative assets---we design a scale prediction module that estimates the scale factor of the input point cloud and rescales it to a canonical human size (\cref{sec:method:scale_pred}). Together, these three main modules constitute the \modelname framework, below we will describe their designs and training details.


\subsection{Preliminary}
\label{sec:method:preli}

\pheading {Parametric Body Model -- \smplx.} \smplx~\cite{SMPLX:2019} has been a
standard body format in various clothed human datasets~\cite{wang20244ddress, tao2021thuman2.1, human4dit, han20232k2k, ho2023customhumans, shen2023xavatar}. It is a statistical model that maps body shape $\shapecoeff \in \real^{10}$, pose $\posecoeff \in \real^{\joints \times 3}$, and facial expression $\expressioncoeff \in \real^{10}$ to mesh vertices $\vsmplx \in \real^{10475 \times 3}$, where $\joints$ is the number of human joints ($\joints = 55$, containing body, eyes, jaw and finger joints in addition to a joint for global rotation). $\shapecoeff$ are linear shape coefficients of the shape blendshapes, and $B_{S}(\shapecoeff)$ accounts for variations of body shapes. $\posecoeff$ contains the relative rotation (axis-angle) of each joint plus the root one \wrt their parent in the kinematic tree, and $B_{P}(\posecoeff)$ models the pose-dependent deformation. $\expressioncoeff$ are PCA coefficients of the expression blend shape function, and $B_{E}(\expressioncoeff)$ accounts for variations of facial expressions. Shape displacements $B_{S}(\shapecoeff)$, pose correctives $B_{P}(\posecoeff)$ and facial expression displacements $B_{E}(\expressioncoeff)$ are added together onto the template mesh $\template \in \real^{10475 \times 3}$, in the rest pose (or T-pose), to produce the output mesh $\restpose$:
\begin{equation}
    \label{eq:smplx_t}\restpose(\shapecoeff, \posecoeff, \expressioncoeff) = \template
    + B_{S}(\shapecoeff) + B_{P}(\posecoeff)+B_{E}(\expressioncoeff),
\end{equation}
Next, the joint regressor $J(\shapecoeff)$ is applied to the rest-pose mesh $\restpose$ to obtain the 3D joints : $\real^{\shapedim}\to \real^{J \times 3}$. Finally, Linear Blend Skinning (LBS) $W(\cdot)$ is used for reposing purposes, the skinning weights are denoted as $\blendweights$, then the posed mesh is translated with $\translcoeff\in \real^{3}$ as final output $\mathbf{M}$ :
\begin{equation}
    \label{eq:smplx_lbs}\mathbf{M}(\shapecoeff, \posecoeff, \expressioncoeff, \translcoeff
    ) = W( \restpose(\shapecoeff, \posecoeff, \expressioncoeff), J(\shapecoeff),
    \posecoeff, \boldsymbol{t}, \blendweights ). 
\end{equation}

\begin{figure*}[t]
    \centering
    \includegraphics[trim=000mm 000mm 000mm 000mm, clip=true, width=\linewidth]{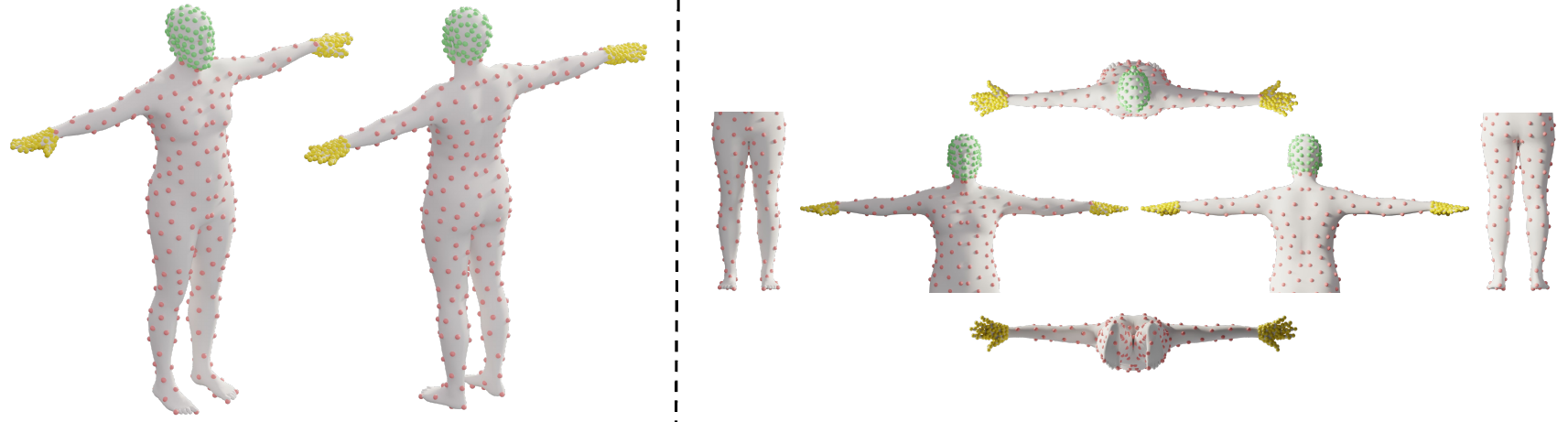}
    \vspace{-4mm}
    \caption{\scriptsize \textbf{Landmark Distribution on the \smplx Template.} We place a total of 600 landmarks across three body regions: \textcolor{HeadGreen}{120 for the head}, \textcolor{BodyRed}{300 for the body}, and \textcolor{HandYellow}{180 for the hands}.}
    \label{fig:lmk_dist}
    \vspace{-1em}
\end{figure*}

\subsection{Dense Landmark Prediction}

\label{sec:method:lmk}
Given a point cloud $\pcds = \{\pcd_{i} \in \real^{3}\}_{i=1}^{\numpcds}$ of $\numpcds$ points, our landmark predictor estimates a set of dense landmarks $\lmks = \{\lmk_{i} \in \real^{3}\}_{i=1}^{\numlmks}$, each of which corresponds to a predefined vertex on the \smplx template mesh. We define $\numlmks = 600$ landmarks in total, comprising 120 head, 300 body, and 180 hand landmarks to provide comprehensive coverage of the human body (see~\cref{fig:lmk_dist}).

\begin{figure*}[t]
    \centering
    \includegraphics[trim=000mm 000mm 000mm 000mm, clip=true, width=\linewidth]{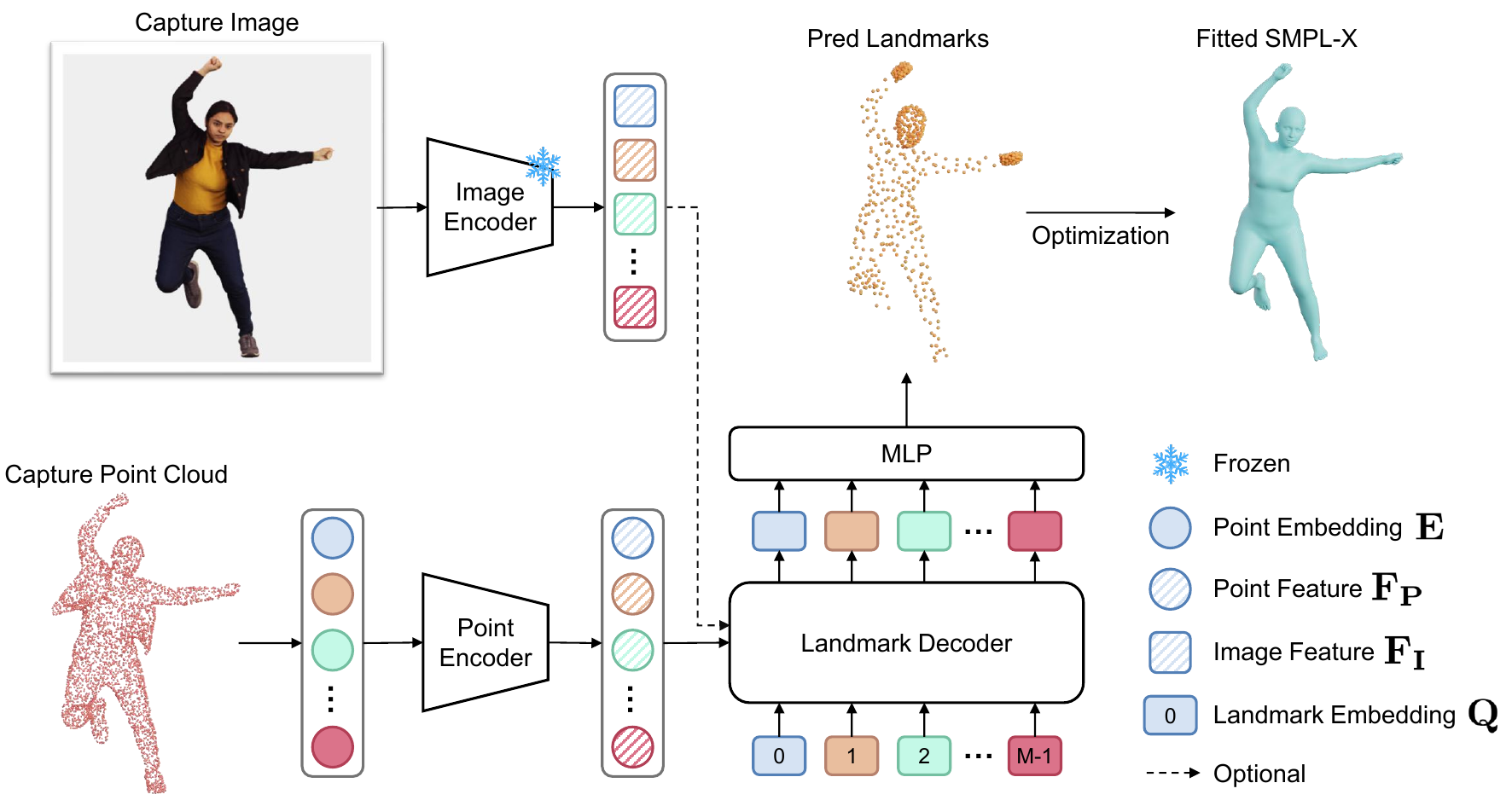}
    \vspace{-4mm}
    \caption{\scriptsize \textbf{The Structure of Landmark Predictor.} Given an input point cloud, the landmark predictor estimates dense 3D landmarks. The point cloud is first tokenized into point embeddings and then fed into the point encoder to produce point features. The landmark decoder is a Perceiver-style transformer that takes learnable landmark embeddings as queries, cross-attends to the point cloud features, and regresses the final 3D landmark coordinates via an MLP. Optionally, image features extracted by a pretrained image encoder can be injected into the landmark decoder, further enhancing the prediction results, which is detailed in~\cref{sec:method:adapter} and~\cref{fig:adapter_scale_predictor}.}
    \label{fig:lmk_predictor}
    \vspace{-2.0em}
\end{figure*}

As shown in~\cref{fig:lmk_predictor}, the landmark predictor follows a conditional decoder architecture with two main components: a point cloud encoder and a landmark decoder. The point cloud encoder uses a Point-BERT~\cite{yu2022pointbert} backbone to tokenize the input into patch embeddings $\pcdembeds = \{\pcdembed_{i}\}_{i=1}^{\numgroup}$, which are then refined by transformer layers to produce per-patch features $\pcdfeats \in \real^{\numgroup \times c}$, where $\numgroup$ is the number of patches and $c$ is the feature dimension. The landmark decoder is a Perceiver-style transformer~\cite{jaegle2021perceiver} that uses $\numlmks$ learnable landmark embeddings $\lmkembeds \in \real^{\numlmks \times c}$ as queries. Each decoder layer applies:

\begin{gather}
    i\in [0, \numlayers], \lmkfeats^{0} = \lmkembeds,\notag\\
    \lmkfeats^{i} = \mathrm{CrossAttn}(\lmkfeats^{i}, \pcdfeats) + \lmkfeats^{i}, \label{eq:lmk_decoder}\\
    \lmkfeats^{i+1} = \mathrm{SelfAttn}(\lmkfeats^{i}) + \lmkfeats^{i}. \notag
\end{gather}    

\noindent $\mathrm{CrossAttn}(\cdot)$ and $\mathrm{SelfAttn}(\cdot)$ denote the cross-attention and self-attention layers. $\numlayers$ indicates the number of transformer layers in the landmark decoder, and $\lmkfeats^{i} \in \real^{\numlmks \times c}$ is the hidden states of the $i$-th layer. After the last layer, a $\mathrm{MLP}$ block is applied to the final landmark features $\lmkfeats^{\numlayers}$ to obtain the 3D coordinates of the predicted landmarks $\lmks = \mathrm{MLP}(\lmkfeats^{\numlayers}) \in \real^{\numlmks \times 3}$.

In~\cref{fig:lmk_predictor}, our landmark predictor can also optionally take an additional condition from extracted image features via a plug-and-play image adapter structure, which is detailed in~\cref{sec:method:adapter}.

\begin{figure*}[t]
    \centering
    \includegraphics[trim=000mm 000mm 000mm 000mm, clip=true, width=\linewidth]{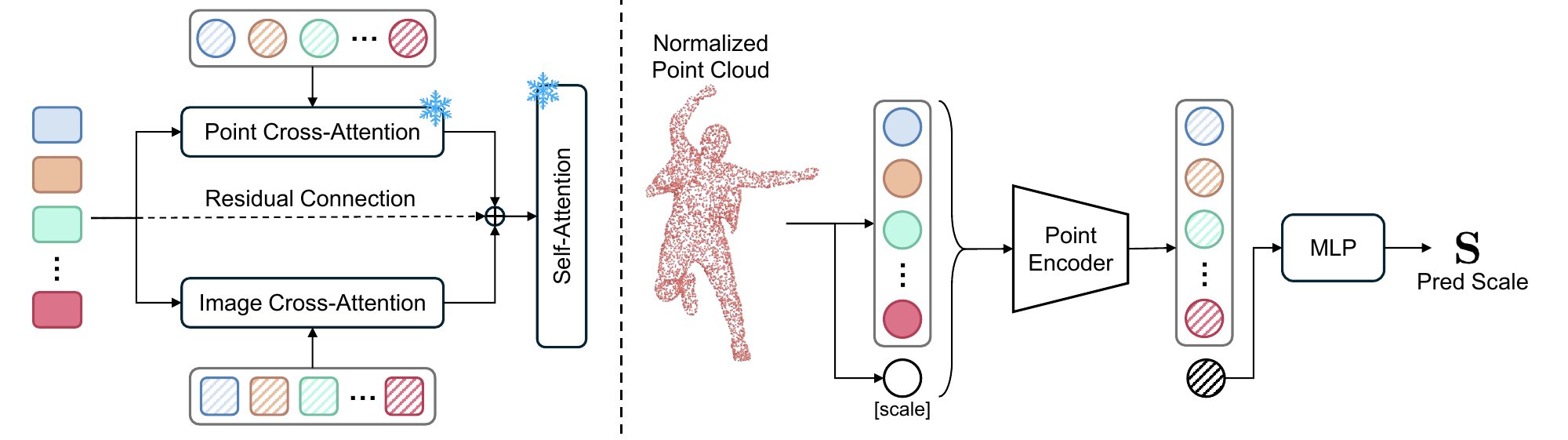}
    \caption{\scriptsize \textbf{Image Adapter (Left) and Scale Predictor (Right).} \textbf{(Left)} The plug-and-play image adapter adds a lightweight cross-attention branch in parallel with the existing point cloud cross-attention in each landmark decoder layer. Image features are fused alongside point cloud features without modifying the base architecture, allowing the adapter to be enabled or disabled at will. \textbf{(Right)} The scale predictor shares the same architecture as the point cloud encoder, but adds scale token to the patch embedding sequence. The token's output is passed through an MLP to regress a scale factor $\scalefactor$, which rescales the input point cloud to canonical human size.}
    \label{fig:adapter_scale_predictor}
    \vspace{-4em}
\end{figure*}

\subsection{Plug-and-Play Image Adapter}
\label{sec:method:adapter}

Images are a common companion to 3D human assets, available as rendered RGB images from 3D meshes or 3DGS~\cite{kerbl20233dgs}, or RGBD captures of real humans. Compared to point cloud alone, images carry richer details---such as facial features, hand poses, and clothing wrinkles---that benefit accurate body fitting.

Inspired by MV-Adapter~\cite{huang2024mvadapter}, we propose a plug-and-play image adapter that takes an optional image $\image$ as an extra input to improve landmark prediction. As shown in~\cref{fig:adapter_scale_predictor}, the adapter adds a lightweight cross-attention branch in parallel with the point cloud cross-attention in each decoder layer, with no change to the base architecture. Image features $\imagefeats$ are extracted by a pretrained image encoder and fused in parallel with the point cloud features. The updated cross-attention in~\cref{eq:lmk_decoder} becomes:

\begin{equation}
    \label{eq:adapter}
    \lmkfeats^{i} = \mathrm{CrossAttn}(\lmkfeats^{i}, \pcdfeats) + \mathrm{CrossAttn}(\lmkfeats^{i}, \imagefeats) + \lmkfeats^{i}.
\end{equation}

This design keeps the base model intact, allowing the adapter to be enabled or disabled at will during both training and inference, making the framework flexible for scenarios where images may or may not be available.

\subsection{Scale Prediction}
\label{sec:method:scale_pred}

3D human assets, particularly those from generative models, often exhibit scale distortion that degrades body fitting accuracy. This arises because generative methods typically normalize outputs to a unit bounding box for training stability, introducing scale ambiguity---for instance, a seated person normalized in the unit box will appear at a larger scale than a canonical standing human.

To address this, we design a scale predictor that takes a normalized point cloud $\normpcds$ of arbitrary pose and estimates a scale factor $\scalefactor$ to bring it to canonical human size. As illustrated in~\cref{fig:adapter_scale_predictor}, the scale predictor shares a similar architecture with the point cloud encoder, but adds a learnable scale token $\scaletoken$ to the input patch embedding sequence: $\{\scaletoken, \normpcdembed_{1}, \normpcdembed_{2}, \dots, \normpcdembed_{g}\}$. After passing through the transformer layers, the output at the scale token position is fed into an MLP to regress $\scalefactor$. The rescaled point cloud $\pcds = \normpcds \cdot \scalefactor$ is then passed to the landmark predictor.

\subsection{Training and \smplx Optimization}
\label{sec:method:train_optim}
The landmark predictor is trained end-to-end using an MSE loss between the predicted landmarks $\lmks = \{\lmk_{i} \in \real^{3}\}_{i=1}^{\numlmks}$ and the ground-truth $\gtlmks = \{\gtlmk_{i} \in \real^{3}\}_{i=1}^{\numlmks}$:

\begin{equation}
    \label{eq:lmk_loss}
    \mathcal{L}_{\mathrm{lmk}} = \frac{1}{\numlmks} \sum_{i=1}^{\numlmks} \|\lmk_{i} - \gtlmk_{i}\|_{2}^{2}.
\end{equation}

\noindent The point cloud encoder and landmark decoder are trained jointly. The image adapter is then attached to the pretrained landmark predictor for adapter training, with the landmark predictor weights frozen and only the adapter branch updated using the same loss in~\cref{eq:lmk_loss}. The scale predictor is trained independently with an MSE loss between the predicted scale factor $\scalefactor$ and the ground-truth $\gtscalefactor$:
\begin{equation}
    \label{eq:scale_loss}
    \mathcal{L}_{\mathrm{scale}} = \|\scalefactor - \gtscalefactor\|_{2}^{2}.
\end{equation}

\noindent At inference, \smplx parameters $\{\shapecoeff, \posecoeff, \expressioncoeff, \translcoeff\}$ are optimized iteratively by minimizing the landmark alignment objective:
\begin{equation}
    \label{eq:optim_smplx}
    \underset{\shapecoeff, \posecoeff, \expressioncoeff, \translcoeff}{\mathrm{argmin}} \sum_{i=1}^{\numlmks} \|\optimlmk_{i} - \lmk_{i}\|_{2}^{2},
\end{equation}
where $\optimlmk_{i}$ denotes the $i$-th landmark on the current \smplx estimate.




\label{sec:method:dataset}

\section{Experiments}
\label{sec:expr}

\subsection{Datasets.}
\label{sec:expr:datasets}

\pheading{Landmark Prediction.}
For benchmark comparison, we follow the dataset setting and splits in \etch~\cite{li2025etch}, using the \cape~\cite{ma2020cape} and \ddress~\cite{wang20244ddress} datasets. For training the generalizable model, we construct a unified dataset by combining \cape, \ddress, and two additional synthetic datasets. Specifically, the synthetic data are prepared as follows: (1) We augment the 3D clothed human assets from the BEDLAM2 dataset~\cite{teschbedlam2} with hair~\cite{he2025perm} and shoes, yielding 98,659 frames; (2) We retarget Motion-X~\cite{lin2023motionx} pose sequences onto the 3D human templates from the SynBody dataset~\cite{yang2023synbody}, resulting in 158,410 frames. The inclusion of synthetic data substantially enriches the diversity of human poses and geometric variations, and critically, provides perfectly accurate ground-truth underlying body models, which are essential for training a generalizable model.

\pheading{Image Adapter.}
For comparison with multi-view image fitting methods, we use the \ddress dataset with the same splits as described above. Since \ddress provides textured 3D human meshes, we render front-view RGB images from these meshes as a condition using orthographic camera projection. We also train a generalizable image adapter with extended training data, including point cloud and front-view rendered image from 2K2K~\cite{han20232k2k} and X-Humans~\cite{shen2023xavatar}.

\pheading{Scale Prediction.}
For scale predictor training, we use the unified dataset in landmark prediction, which includes \cape, \ddress, and the two synthetic datasets. The input point cloud is normalized to $[-0.9,0.9]$, and the scale predictor is trained to predict the scale factor that can restore the original size of the input.

\pheading{Partial Data.}
Considering the practical scenarios where only partial data may be available, we simulate the most common partial data pattern, single-view, \ie, from a certain view angle, only the front part is visible. Given a full mesh, to simulate the single-view point cloud, we remove the vertices and faces not visible from the front view, and then sample points from the remaining mesh surface.

\subsection{Metrics}
\label{sec:expr:metrics}
We evaluate our method using two metrics: (1) \textbf{Vertex-to-Vertex (V2V) distance} (cm), which measures the mean Euclidean distance between the predicted body mesh vertices and the corresponding ground-truth vertices; and (2) \textbf{Mean Per Joint Position Error (MPJPE)} (cm), which measures the mean Euclidean error over $\joints$ \smplx body joints. In all reported tables, lower values indicate more accurate body fitting results.

\subsection{Implementation Details}
\label{sec:expr:impl}
The landmark decoder consists of $\numlayers=24$ transformer blocks, each comprising self-attention and cross-attention layers. Both the point cloud encoder and the scale predictor adopt a PointBERT~\cite{yu2022pointbert} architecture with 16 and 12 transformer blocks, respectively. The image encoder is DINOv2~\cite{oquab2024dinov2}. All models are trained on the target dataset using the AdamW optimizer with a learning rate of $5 \times 10^{-5}$ for 100 epochs. The per-GPU batch sizes are set to 20, 24, and 64 for the landmark predictor, image adapter, and scale predictor, respectively. Training is conducted on 8 NVIDIA 48GB GPUs. The number of landmarks is fixed at $\numlmks = 600$. During training, the number of input points is randomly sampled from $[5000, 20000]$ with random rotations applied for augmentation, and $50\%$ of the full point clouds are replaced with partial ones to improve robustness to inputs.

Based on~\cite{li2025etch}, we apply a three-stage optimization procedure after landmark prediction: (1) optimizing translation $\translcoeff$ for 20 steps with $\text{lr} = 5 \times 10^{-1}$; (2) jointly optimizing shape coefficients $\shapecoeff[:2]$ and pose coefficients $\posecoeff$ for 30 steps with $\text{lr} = 5 \times 10^{-1}$; and (3) optimizing all parameters for 20 steps with $\text{lr} = 2 \times 10^{-1}$.

\begin{table*}[t]
    \centering
    \caption{\scriptsize \textbf{Comparison with Point Cloud Body Fitting Methods.} Our approach outperforms all baselines by a large margin on both \cape and \ddress, and is the first method to achieve millimeter-level body fitting accuracy. Compared to previous \sota \etch~\cite{li2025etch}, \modelname reduces V2V error by \textcolor{GreenColor}{57.1\%}/\textcolor{GreenColor}{75.5\%} and MPJPE by \textcolor{GreenColor}{67.1\%}/\textcolor{GreenColor}{80.9\%} on \cape/\ddress. The $\Delta$ row reports the per-region relative improvement over the second-based method, with particularly pronounced gains on the hands and head, benefiting from our dense landmark distribution.}
    \renewcommand{\arraystretch}{1.2}
    \resizebox{\textwidth}{!}
    {
        \begin{tabular}{c|cccc|cccc|cccc|cccc}
       \bottomrule
        \multirow{3}{*}{Methods} & \multicolumn{8}{c|}{\cape}             & \multicolumn{8}{c}{\ddress}              \\ \cline{2-17}
                                 & \multicolumn{4}{c|}{V2V $\downarrow$}  & \multicolumn{4}{c|}{MPJPE $\downarrow$}  & \multicolumn{4}{c|}{V2V $\downarrow$}  & \multicolumn{4}{c}{MPJPE $\downarrow$} \\ \cline{2-17}
                                 & All    & Hands  & Head   & Body        & All    & Hands  & Head   & Body          & All             & Hands           & Head            & Body                 & All             & Hands           & Head            & Body         \\ \shline
        \ipnet                   & 5.529  & 7.454  & 5.485  & 5.001       & 5.611  & 6.600  & 4.527  & 4.399         & 7.495           & 8.881           & 7.378           & 7.178                & 7.380           & 8.606           & 5.973           & 5.894        \\  
        \ptf                     & 2.341  & 3.880  & 2.038  & 2.099       & 2.641  & 3.377  & 1.720  & 1.767         & 3.297           & 4.938           & 3.338           & 2.785                & 3.567           & 4.607           & 2.612           & 2.248        \\
        \nicp                    & 1.736  & 2.741  & 1.184  & 1.827       & 2.074  & 2.565  & 1.042  & 1.597         & 4.085           & 6.224           & 3.323           & 3.993                & 4.862           & 6.142           & 2.540           & 3.521        \\
        \arteq                   & 2.202  & 3.417  & 2.011  & 1.943       & 2.405  & 3.055  & 1.693  & 1.589         & 3.072           & 4.537           & 3.145           & 2.636                & 3.378           & 4.170           & 2.335           & 2.156        \\
        \etch                    & 1.567  & 3.449  & 1.236  & 1.240       & 2.002  & 2.833  & 0.928  & 1.007         & 2.408           & 5.108           & 1.997           & 2.178                & 3.459           & 4.695           & 1.420           & 2.141        \\ 
        Ours                     & \textbf{0.672}  & \textbf{0.670} & \textbf{0.492} & \textbf{0.798}      & \textbf{0.659}  & \textbf{0.670} & \textbf{0.469}  & \textbf{0.689}       & \textbf{0.589}  & \textbf{0.834}  & \textbf{0.511}  & \textbf{0.570}  & \textbf{0.662}   & \textbf{0.782}  &  \textbf{0.433} &  \textbf{0.539}\\ \hline
      $\Delta$  & \textcolor{GreenColor}{57.1\%} & \textcolor{GreenColor}{80.6\%} & \textcolor{GreenColor}{60.2\%} & \textcolor{GreenColor}{35.6\%} & \textcolor{GreenColor}{67.1\%} & \textcolor{GreenColor}{76.4\%} & \textcolor{GreenColor}{49.5\%} & \textcolor{GreenColor}{31.6\%} & \textcolor{GreenColor}{75.5\%} & \textcolor{GreenColor}{83.7\%} & \textcolor{GreenColor}{74.4\%} & \textcolor{GreenColor}{78.4\%} & \textcolor{GreenColor}{80.9\%} & \textcolor{GreenColor}{83.3\%} & \textcolor{GreenColor}{69.5\%} & \textcolor{GreenColor}{74.8\%} \\
       
       \toprule
        \end{tabular}
    
    }

    \label{tab:comparison}
    \vspace{-1em}
\end{table*}

\begin{figure*}[t]
    \centering
    \vspace{-1.0em}
    \includegraphics[trim=000mm 000mm 000mm 000mm, clip=true, width=\linewidth]{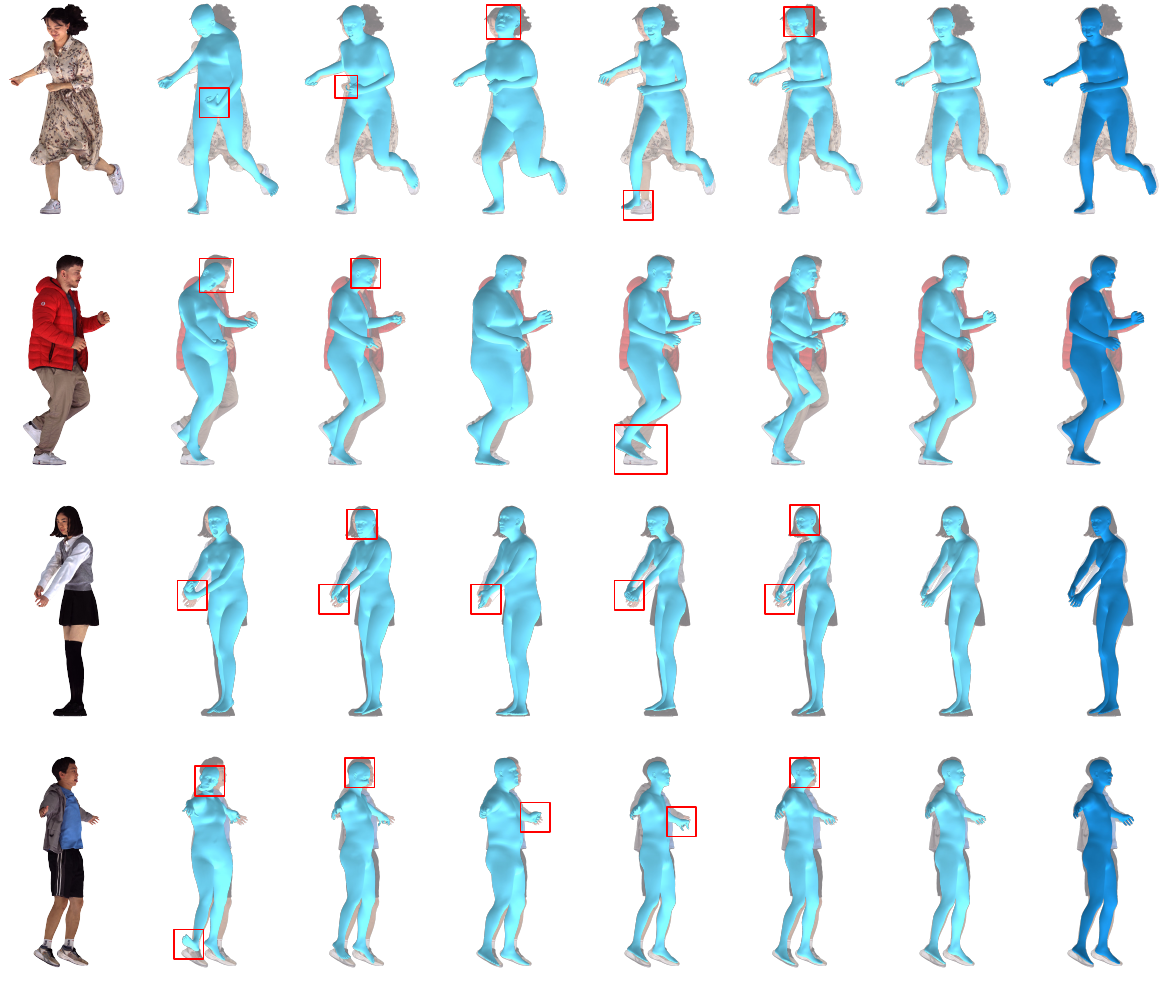}

    \begin{tabularx}{\linewidth}{
        @{}p{0.002\linewidth}@{}
        >{\centering\arraybackslash}p{0.118\linewidth}
        >{\centering\arraybackslash}p{0.118\linewidth}
        >{\centering\arraybackslash}p{0.118\linewidth}
        >{\centering\arraybackslash}p{0.118\linewidth}
        >{\centering\arraybackslash}p{0.118\linewidth}
        >{\centering\arraybackslash}p{0.118\linewidth}
        >{\centering\arraybackslash}p{0.118\linewidth}
        >{\centering\arraybackslash}p{0.118\linewidth}
        } 
        & \scriptsize 3D Human & \scriptsize \ipnet~\cite{bhatnagar2020ipnet} & \scriptsize \ptf~\cite{wang2021ptf} & \scriptsize \nicp~\cite{marin24nicp} & \scriptsize \arteq~\cite{feng2023arteq} & \scriptsize \etch~\cite{li2025etch} & \scriptsize Ours & \scriptsize GT\\
    \end{tabularx}
    
    \vspace{-1.0em}
    \caption{\scriptsize \textbf{Qualitative Comparison with Point Cloud Body Fitting Methods.} \modelname produces more accurate \smplx fitting compared to all baselines, especially in the hand and head regions. Failure cases of competing methods are highlighted with \textcolor{red}{red boxes}, showing common artifacts such as incorrect hand poses, misaligned head orientations, and inaccurate body shapes. In contrast, \modelname consistently recovers fine-grained body details across all cases. Please zoom in for a better view.}
    \label{fig:quatitative}
    \vspace{-2.0em}
\end{figure*}
\subsection{Point Cloud Fitting Comparison}
\label{sec:expr:pcd_fit}
With only point cloud input, we compare out method, \modelname, with multiple \sota point cloud body fitting baselines~\cite{bhatnagar2020ipnet,wang2021ptf,marin24nicp,feng2023arteq,li2025etch}, as shown in~\cref{tab:comparison}. Note that \modelname predicts \smplx bodies while previous methods only predict \smpl or \smplh bodies. For a fair comparison, we implement the \smplx version of these methods, and all the results are evaluated on the \smplx body model with the same train and test settings.

Overall, \modelname achieves the best performance across all datasets and metrics, and is the first method to reduce body fitting errors to the millimeter level. In particular, on CAPE, our approach outperforms the second-best method, \etch~\cite{li2025etch}, by a significant reduction of \textcolor{GreenColor}{57.1\%} in V2V error and \textcolor{GreenColor}{67.1\%} in MPJPE; on \ddress, the improvement is even more significant with a \textcolor{GreenColor}{75.5\%} reduction in V2V error and \textcolor{GreenColor}{80.9\%} reduction in MPJPE. These results demonstrate the effectiveness of our conditional landmark decoder design.

The last row of~\cref{tab:comparison} represents the percentage improvement of our method compared to the second-best method, and it is worth noting that \modelname has better performance, especially on hands and head regions. We attribute this to the flexible conditional decoder design of landmark predictor, which allows us to adaptively allocate more landmarks to more complex regions, thus the training target will capture finer details and improving final fitting accuracy. \Cref{fig:quatitative} shows the visual comparison of the body fitting results, where we can see that \modelname produces more accurate \smplx fitting compared to the baselines, especially in detailed regions like hands and head.

\begin{table}[t]
    \centering
    \begin{minipage}[t]{0.35\textwidth}
        \centering
        \caption{\scriptsize \textbf{Comparison with Multi-View Body Fitting.} \modelname (point cloud only) already surpasses multi-view methods, and the image adapter further improves accuracy.}
        \vspace{-1em}
        \renewcommand{\arraystretch}{1.2}
        \resizebox{0.9\textwidth}{!}{
            \begin{tabular}{c|cc}
                \bottomrule
                & V2V $\downarrow$ & MPJPE $\downarrow$ \\
                \hline
                EasyMocap & 3.047 & 4.029 \\
                DiffProxy & 2.093 & 2.630 \\
                \hline
                Ours & 0.589 & 0.662 \\
                Adapter & \textbf{0.539} & \textbf{0.622} \\
                \toprule
            \end{tabular}
        }
        \label{tab:image_adapter_compare}
    \end{minipage} \hspace{0.03\textwidth}
    \begin{minipage}[t]{0.60\textwidth}
        \centering
        \caption{\scriptsize \textbf{Unified Dataset vs. \ddress.} Training on large-scale unified data (*) improves generalization to out-of-distribution datasets (CustomHumans, THuman2.1) while maintaining competitive performance on \ddress.}
        \vspace{-1.2em}
        \renewcommand{\arraystretch}{1.2}
        \resizebox{\textwidth}{!}{
            \begin{tabular}{c|cc|cc|cc}
                \bottomrule
                \multirow{2}{*}{Methods} & \multicolumn{2}{c|}{\ddress} & \multicolumn{2}{c|}{CustomHumans} & \multicolumn{2}{c}{THuman2.1} \\ \cline{2-7}
                & V2V$\downarrow$ & MPJPE$\downarrow$ & V2V$\downarrow$ & MPJPE$\downarrow$ & V2V$\downarrow$ & MPJPE$\downarrow$ \\ \hline
                Ours     & 0.589 & 0.662 & 1.187 & 1.280 & 1.768 & 1.811 \\
                Ours*    & 0.580 & 0.643 & 1.112 & 1.117 & 1.338 & 1.524 \\ \hline
                Adapter  & 0.539 & 0.622 & 1.181 & 1.278 & 1.767 & 1.806 \\
                Adapter* & 0.526 & 0.610 & 0.878 & 0.973 & 0.835 & 1.144 \\
                \toprule
            \end{tabular}
        }
        \label{tab:compare_unified}
    \end{minipage}
    \vspace{-1em}
\end{table}

\begin{figure*}[t]
    \centering
    \includegraphics[trim=000mm 000mm 000mm 000mm, clip=true, width=\linewidth]{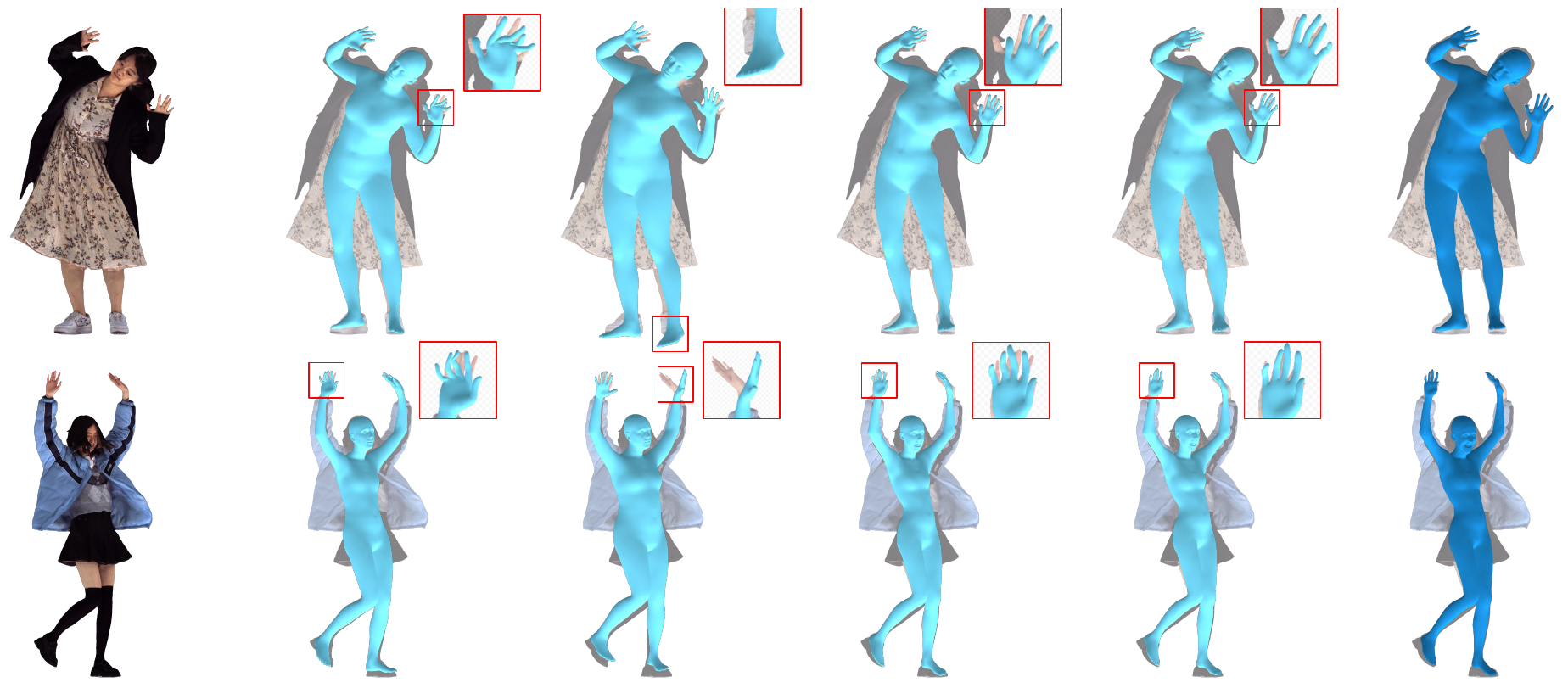}
    
    \begin{tabularx}{\linewidth}{
        @{}p{0\linewidth}@{}
        >{\centering\arraybackslash}p{0.125\linewidth}
        >{\centering\arraybackslash}p{0.188\linewidth}
        >{\centering\arraybackslash}p{0.158\linewidth}
        >{\centering\arraybackslash}p{0.168\linewidth}
        >{\centering\arraybackslash}p{0.168\linewidth}
        >{\centering\arraybackslash}p{0.168\linewidth}
        } 
        & \scriptsize 3D Human & \scriptsize DiffProxy~\cite{wang2025diffproxy} & \scriptsize EasyMocap~\cite{easymocap} & \scriptsize Ours & \scriptsize Adapter & \scriptsize GT\\
    \end{tabularx}
    \vspace{-1.5em}
    \caption{\scriptsize \textbf{Comparison with Multi-View Body Fitting.} \modelname with the image adapter outperforms multi-view methods DiffProxy~\cite{wang2025diffproxy} and EasyMocap~\cite{easymocap} in body fitting accuracy. Details are highlighted with \textcolor{red}{red boxes} and magnified, showing that competing methods struggle with fine-grained regions such as hands, while our adapter recovers these details by incorporating point cloud and image.}
    \label{fig:adapter_comparison}
    \vspace{-1.5em}
\end{figure*}

\subsection{Image Adapter}
\label{sec:expr:adapter}
\vspace{-1.0 em}
The image adapter enables \modelname to incorporate RGB images as additional input, further enhancing fitting performance. We evaluate \modelname with and without the image adapter on the \ddress dataset, and also compare against \sota multi-view image fitting methods~\cite{easymocap,wang2025diffproxy} that use multi-view RGB images (with camera pose) as input. Throughout all tables and figures, \textbf{Adapter} denotes the use of the image adapter.

As shown in \cref{tab:image_adapter_compare}, with the image adapter, \modelname not only surpasses its point cloud-only backbone but also outperforms leading multi-view image fitting methods. These results demonstrate that point clouds provide a strong signal for 3D body fitting, while the addition of image information further improves accuracy by supplying fine-grained details. The visualization results in~\cref{fig:adapter_comparison} further confirm this, additional image condition making detailed hands region fitting better.  By effectively combining point cloud and image cues, the image adapter enables \modelname to achieve superior fitting results compared to approaches relying solely on multi-view images.

\begin{table}[t]
    \centering
    \begin{minipage}[t]{0.3\textwidth}
        \centering
        \caption{\scriptsize \textbf{Effectiveness of Scale Predictor.} For scale-normalized 3D humans, incorporating scale predictor improves body fitting accuracy.}
        \vspace{-1em}
        \renewcommand{\arraystretch}{1.2}
        \resizebox{0.95\textwidth}{!}{
            \begin{tabular}{c|cc}
                \bottomrule
                & V2V$\downarrow$ & MPJPE$\downarrow$ \\
                \hline
                gt scale & 0.589 & 0.662 \\
                w/o scale pred & 1.120 & 1.172 \\
                w/ scale pred & 0.651 & 0.709 \\
                \toprule
            \end{tabular}
        }
        \label{tab:scale_pred}
    \end{minipage}%
    \hfill%
    \begin{minipage}[t]{0.68\textwidth}
        \centering
        \caption{\scriptsize \textbf{Ablation on Landmark Distribution.} Given a total number of 600 landmarks, we evaluate different allocations across body parts and find that setting C. achieves the best overall performance.}
        \vspace{-1em}
        \renewcommand{\arraystretch}{1.2}
        \resizebox{\textwidth}{!}
        {
           \begin{tabular}{c|ccc|cccc|cccc}
           \bottomrule
                     &  \multicolumn{3}{c|}{Landmarks} & \multicolumn{4}{c|}{V2V $\downarrow$} & \multicolumn{4}{c}{MPJPE $\downarrow$} \\ \cline{1-12}
           Settings  &  Hands  &   Head   &  Body   &  All   &  Hands  & Head   & Body   & All    & Hands  &  Head  &   Body \\ \cline{1-12}
           A.        &  300    &   120    &  180    &  0.655 &  \textbf{0.832}  & 0.577  & 0.657  & 0.701  & 0.793  &  0.499 &  0.614  \\ 
           B.        &  240    &   120    &  240    &  0.621 &  0.833  & 0.524  & 0.626  & 0.690  & 0.792  &  0.445 &  0.597  \\
           \rowcolor{gray!20}
           C.        &  180    &   120    &  300    &  \textbf{0.589}  & 0.834  & \textbf{0.511}  & \textbf{0.570}  & \textbf{0.662}   & \textbf{0.782}  &  \textbf{0.433} &  \textbf{0.539}   \\ 
           D.        &  120    &   120    &  360    &  0.637 &  0.955  & 0.550  & 0.603  & 0.724  & 0.875  & 0.477  & 0.558 \\
           
           \toprule
            \end{tabular}
        }
        \label{tab:ablation:lmk_dist}
    \end{minipage}
    \vspace{-0.5em}
\end{table}

\begin{figure*}[t]
    \centering
    \includegraphics[trim=000mm 000mm 000mm 000mm, clip=true, width=\linewidth]{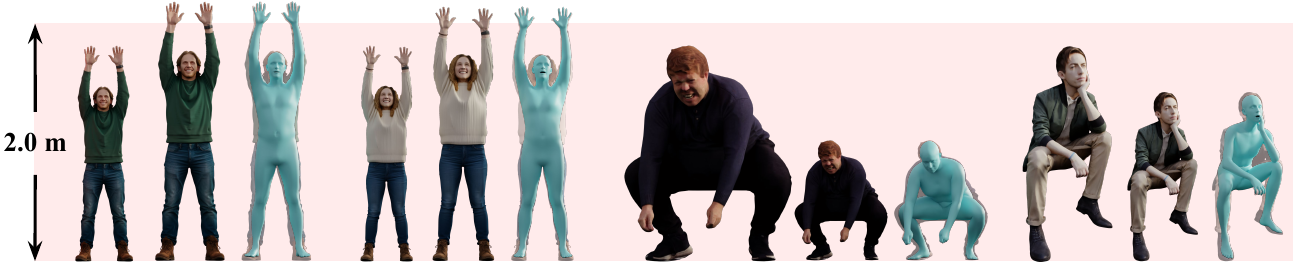}
    \vspace{-2em}
    \caption{\scriptsize \textbf{Rescaling and Body Fitting Results on Generated 3D Humans.} For each example, we show (from left to right) the original 3D human, the rescaled 3D human, and the corresponding \smplx fitting result. \modelname also performs well on scale-distorted human assets.}
    \label{fig:aigc_rescale_fitting_results}
    \vspace{-1.5em}
\end{figure*}

\subsection{Generalizable Model}
To improve the generalization capability of \modelname, we train a generalizable landmark predictor and image adapter on the unified dataset described in~\cref{sec:expr:datasets}. In~\cref{tab:compare_unified}, we compare the generalizable model with our base model trained only on \ddress, and evaluate their performance on the test sets of \ddress and two out-of-distribution datasets, THuman2.1~\cite{tao2021thuman2.1} and CustomHumans~\cite{ho2023customhumans}. Through training on large-scale datasets, our model achieves better body fitting results for 3D humans of different styles.

\subsection{Scale Prediction}
In~\cref{tab:scale_pred}, we demonstrate the effectiveness of the scale predictor. Using the landmark predictor trained on \ddress for body fitting, we normalize the input point cloud to $[-0.9, 0.9]$ to simulate the scale setting of 3D generation methods, and subsequently apply the predicted scale factor to restore the original size of the input point cloud. The results show that the scale predictor reduces the V2V error by \textcolor{GreenColor}{41.9\%} and the MPJPE by \textcolor{GreenColor}{39.5\%}, substantially outperforming the variant without the scale predictor. Combined with the generalizable landmark predictor, \modelname is capable of processing diverse 3D humans produced by generative models. As illustrated in~\cref{fig:aigc_rescale_fitting_results}, our scale predictor first rescales the generated 3D human to a canonical body size and then performing \smplx body fitting, yielding accurate and robust results for 3D humans generated by~\cite{chen2025synchuman}.

\subsection{Ablation Studies}
\begin{figure*}[t]
    \centering
    \includegraphics[trim=000mm 000mm 000mm 000mm, clip=true, width=\linewidth]{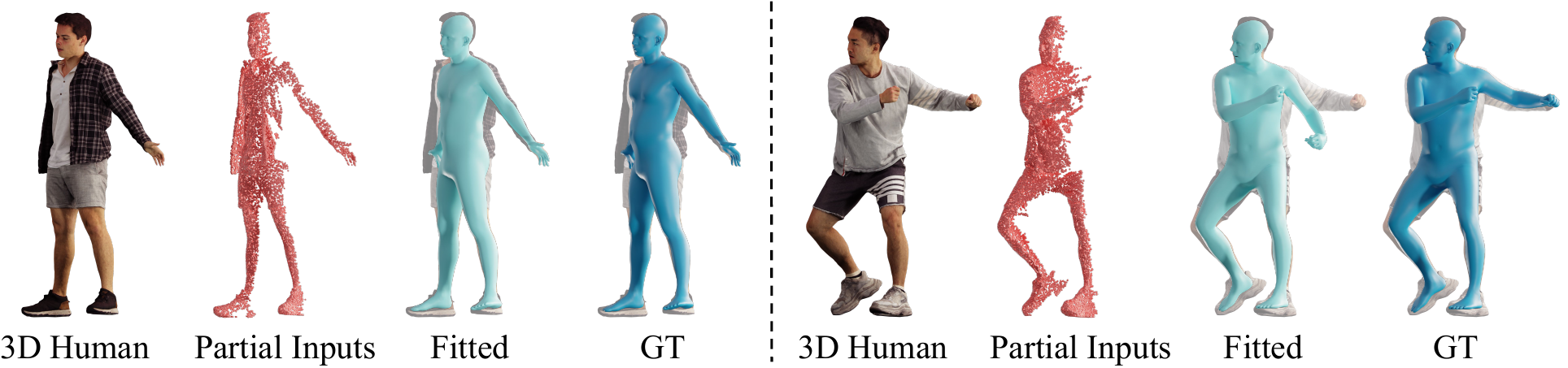}
    \caption{\scriptsize \textbf{Body Fitting Results with Partial Point Cloud Input.} We evaluate \modelname on partial point cloud inputs. In the right-side case, where entire body regions are missing, \modelname still produces reasonable \smplx fitting results, demonstrating its robustness to incomplete inputs.}
    \label{fig:partial_inputs}
\end{figure*}

We conduct ablation studies to explore the impact of different design choices in \modelname. For the landmark predictor, we analyze the effect of landmark settings (\cref{tab:ablation:lmk_dist}), number of landmarks (\cref{tab:ablation:num_lmks}), and number of input points (\cref{tab:ablation:num_pcds}). For the image adapter, we discuss the choice of image encoder (\cref{tab:ablation:image_encoder}). We also evaluate the performance of \modelname with partial inputs (\cref{tab:ablation:partial} and~\cref{fig:partial_inputs}). We finally visualize the point cross-attention maps in the landmark decoder to verify its part correspondence learning process.

\pheading{Landmark Setting.}
As described in~\cref{sec:method:lmk} and illustrated in~\cref{fig:lmk_dist}, we allocate 600 landmarks in total, distributed as 120 for the head, 300 for the body, and 180 for the hands. To validate this design choice, we compare it against alternative allocation settings. As shown in~\cref{tab:ablation:lmk_dist}, our setting achieves the best overall performance, yielding accurate fitting results across all body regions.

\begin{table}[t]
    \centering
    \begin{minipage}[t]{0.4\textwidth}
        \centering
        \caption{\scriptsize \textbf{Ablation on Number of Landmarks.} More landmarks improve fitting accuracy at the cost of higher computation. We select 600 as a trade-off between accuracy and efficiency.}
        \vspace{-1em}
        \renewcommand{\arraystretch}{1.2}
        \resizebox{\textwidth}{!}
        {
            \begin{tabular}{c|cccc}
           \bottomrule
           Num Lmks & V2V$\downarrow$    & MPJPE$\downarrow$   & GFLOPs$\downarrow$   & Iters$\uparrow$\\ \cline{1-5}
           100      & 0.637  & 0.790    & \textbf{23.82}     & \textbf{0.88}      \\
           300      & 0.603  & 0.738    & 51.65     & 0.68      \\
           600      & \textbf{0.589} & \textbf{0.662} & 93.38 & 0.46 \\

           \toprule
            \end{tabular}
        }
        \label{tab:ablation:num_lmks}
    \end{minipage}\hfill
    \begin{minipage}[t]{0.58\textwidth}
       \centering
       \caption{\scriptsize \textbf{Ablation on Number of Input Points.} Fitting accuracy improves steadily across all the body parts with more input points, demonstrating the scalability of \modelname to denser point clouds.}
       \vspace{-1.2em}
        \renewcommand{\arraystretch}{1.2}
        \resizebox{\textwidth}{!}
        {
            \begin{tabular}{c|cccc|cccc}
           \bottomrule
           \multirow{2}{*}{Num PCDs} & \multicolumn{4}{c|}{V2V $\downarrow$} & \multicolumn{4}{c}{MPJPE $\downarrow$} \\ \cline{2-9}
                  &  All   &  Hands  & Head   & Body   & All    & Hands  &  Head  &   Body  \\ \cline{1-9}
           5000   &  0.589 & 0.834   & 0.511  & 0.570  & 0.662  & 0.782  & 0.433  &  0.539  \\ 
           10000  &  0.464 & 0.676   & 0.373  & 0.465  & 0.531  & 0.627  & 0.317  &  0.442  \\ 
           15000  &  \textbf{0.452} & \textbf{0.651}   & \textbf{0.356}  & \textbf{0.460}  & \textbf{0.516}  & \textbf{0.603}  & \textbf{0.300}  &  \textbf{0.438}  \\
           
           \toprule
            \end{tabular}
        }
        \label{tab:ablation:num_pcds}
    \end{minipage}
\end{table}

\begin{table}[t]
    \centering
    \begin{minipage}[t]{0.6\textwidth}
        \centering
        \caption{ \scriptsize \textbf{Ablation on Various Image Encoder Choices.} DINOv2 outperforms Sapiens across all body regions, confirming its effectiveness as the image encoder.}
        \vspace{-1.05em}
        \renewcommand{\arraystretch}{1.2}
        \resizebox{\textwidth}{!}{
            \begin{tabular}{c|cccc|cccc}
                \bottomrule
                \multirow{2}{*}{Encoder} & \multicolumn{4}{c|}{V2V $\downarrow$} & \multicolumn{4}{c}{MPJPE $\downarrow$} \\ \cline{2-9}
                        &  All   &  Hands  & Head   & Body   & All    & Hands  &  Head  &   Body  \\ 
                \cline{1-9}
                Sapiens & 0.589  &  0.836  & 0.513  & 0.570  & 0.661  & 0.783  & 0.436  &  0.533  \\ 
                DINOv2  & \textbf{0.539} & \textbf{0.805} & \textbf{0.449} & \textbf{0.524} & \textbf{0.622} & \textbf{0.751} & \textbf{0.383} & \textbf{0.490} \\ 
                \toprule
            \end{tabular}
        }
        \label{tab:ablation:image_encoder}
    \end{minipage}\hfill
    \begin{minipage}[t]{0.38\textwidth}
       \centering
        \caption{\scriptsize \textbf{Results on Partial Inputs.} \modelname maintains competitive fitting accuracy under partial point cloud.}
        \vspace{-1em}
        \renewcommand{\arraystretch}{1.2}
        \resizebox{\textwidth}{!}
        {
            \begin{tabular}{c|cc|cc}
           \bottomrule
           \multirow{2}{*}{Test} & \multicolumn{2}{c|}{\cape} & \multicolumn{2}{c}{\ddress} \\ \cline{2-5} 
           & V2V$\downarrow$ & MPJPE$\downarrow$ & V2V$\downarrow$ & MPJPE$\downarrow$ \\  \cline{1-5}
           Full    & 0.672 & 0.659 & 0.589 & 0.662 \\
           Partial & 0.887 & 1.001 & 0.595 & 0.671 \\
           \toprule
            \end{tabular}
        }
        
        \label{tab:ablation:partial}
    \end{minipage}
    \vspace{-1.5em}
\end{table}

\pheading{Number of Landmarks.}
Having determined the landmark distribution, we further investigate the effect of the total landmark count. As shown in~\cref{tab:ablation:num_lmks}, increasing the number of landmarks consistently improves fitting accuracy, but also raises the computational cost (GFLOPs) and reduces training throughput (Iters: training iterations per second). We select 600 landmarks as a favorable trade-off between accuracy and efficiency.

\pheading{Number of Input Points.}
For fair comparison with other methods in~\cref{tab:comparison}, we use 5{,}000 input points uniformly. However, \modelname is flexible in the number of input points it can handle. As shown in~\cref{tab:ablation:num_pcds}, fitting performance improves steadily with more input points, demonstrating the scalability of our approach.

\pheading{Choice of Image Encoder.}
We choose DINOv2 as the image encoder in the plug-and-play image adapter, as it has been widely used in human-related perception tasks~\cite{wang2025prompthmr, cai2025up2you, ornek2024foundpose}. In~\cref{tab:ablation:image_encoder}, we compare DINOv2 with Sapiens~\cite{khirodkar2024sapiens}, which is a recent image encoder designed for human-centric tasks. The results show that DINOv2 performs better than Sapiens, indicating that it is more effective in extracting image features for body fitting.

\begin{figure*}[t]
    \centering
    \includegraphics[trim=000mm 000mm 000mm 000mm, clip=true, width=\linewidth]{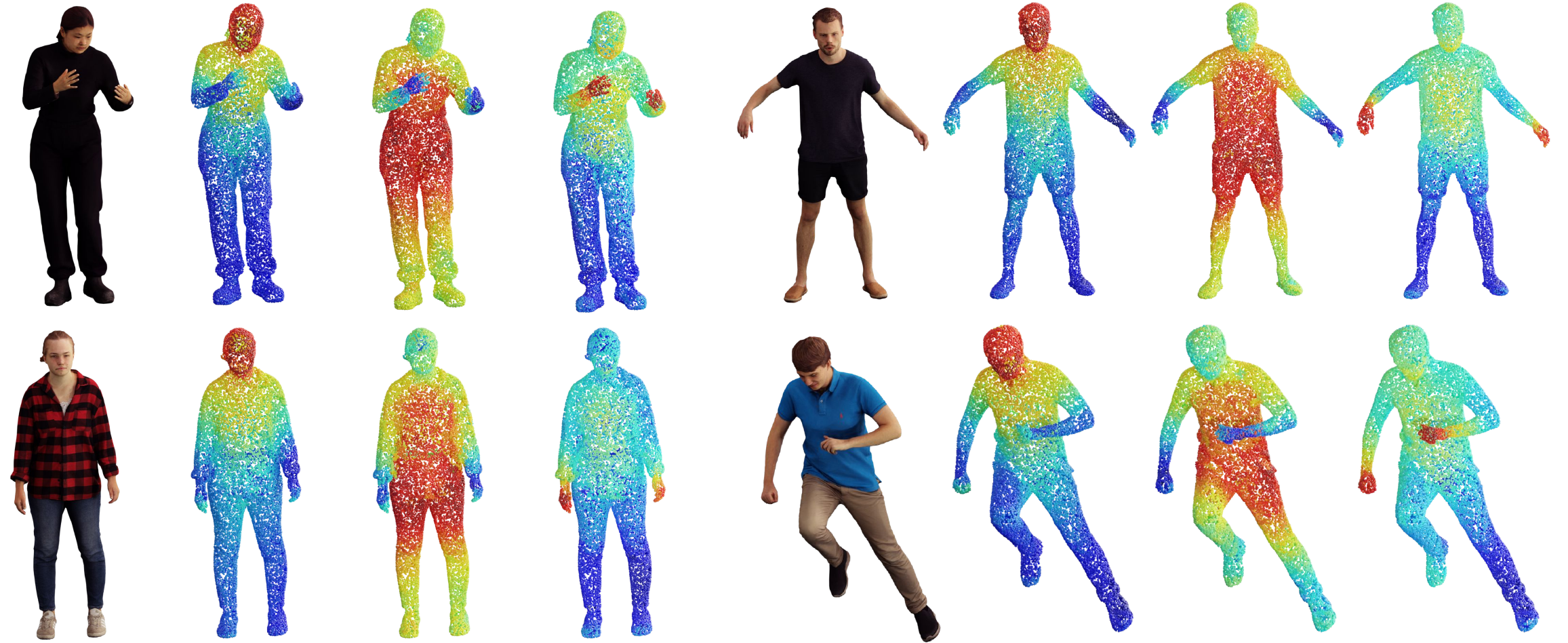}
    \caption{\scriptsize \textbf{Attention Visualization of Landmarks.} We visualize the point cross-attention maps of the landmark decoder. The attention maps accurately reflect part-level correspondence between input surface points and their associated landmarks (Head, Body, Hand).}
    \label{fig:attn_vis}
\end{figure*}
\pheading{Partial Inputs.}
As detailed in~\cref{sec:expr:datasets} (Partial Data), we simulate single-view partial point clouds and incorporate them as supplementary training data alongside full points. Based on this data augmentation, \modelname retains competitive \smplx fitting accuracy under partial inputs, with only an acceptable drop compared to full input (\cref{tab:ablation:partial}). \Cref{fig:partial_inputs} further provides qualitative results, showing that even when an entire body part is missing, our method still produces reasonable body fitting estimates.

\pheading{Part Correspondence Learning in Attention.}
Establishing correspondence between surface points and the template body is essential for 3D body fitting~\cite{bhatnagar2020ipnet,wang2021ptf,li2025etch}. Although our landmark predictor is designed as a conditional decoder that takes the input point cloud as a condition to predict a set of predefined landmarks, its cross-attention maps implicitly capture the correspondence between input surface points and body landmarks. In~\cref{fig:attn_vis}, we visualize the attention maps of the point cross-attention layers in the landmark predictor, with landmarks grouped by head, body, and hand regions. The results demonstrate that the attention maps accurately reflect part-level correspondence, effectively assigning input points to their respective regions on the template body.

\section{Conclusion}
\label{sec:conclusion}

We present \modelname, a unified multi-modal framework for \smplx body fitting across diverse types of 3D human assets.
\modelname combines a conditional transformer decoder for dense landmark prediction, a plug-and-play image adapter for optional RGB fusion, and a scale predictor for resolving scale ambiguity in AI-generated assets.
It is the first body fitting method to achieve \textbf{millimeter-level} accuracy, significantly outperforming prior methods.
Despite these advances, \modelname has two limitations.
First, \smplx parameters are optimized iteratively from predicted landmarks rather than direct regression, which introduces additional inference time.
Second, the image adapter currently supports only a single front-view image as input; extending it to multi-view images could further improve fitting accuracy and robustness. We will leave these as future work to further enhance \modelname's performance and applicability.

\newpage

%
%
\bibliographystyle{splncs04}
\bibliography{main}

\clearpage
\newpage
\appendix
\begin{subappendices}
\renewcommand{\thesection}{\Alph{section}}%
\label{sec:appendix}

\section{Details of Unified Dataset}
To train our generalizable landmark predictor, we construct a unified dataset by combining \cape, \ddress, and two newly collected synthetic datasets. Below, we describe the construction process of each synthetic dataset in detail.

The first synthetic dataset is built based on BEDLAM2~\cite{teschbedlam2}, which provides inner \smplx body meshes with diverse poses and shapes, along with body-aligned clothing. However, BEDLAM2 does not include aligned hair assets, and shoes are represented only as normal and displacement maps without explicit 3D geometry. To obtain more complete and realistic 3D human assets, we supplement these by collecting hair assets from Hair20K~\cite{he2025perm} and 20 shoe assets from the internet, and align them to the clothed body meshes, yielding a total of 98,659 3D human assets.

\begin{figure*}[h]
    \centering
    \includegraphics[trim=000mm 000mm 000mm 000mm, clip=true, width=\linewidth]{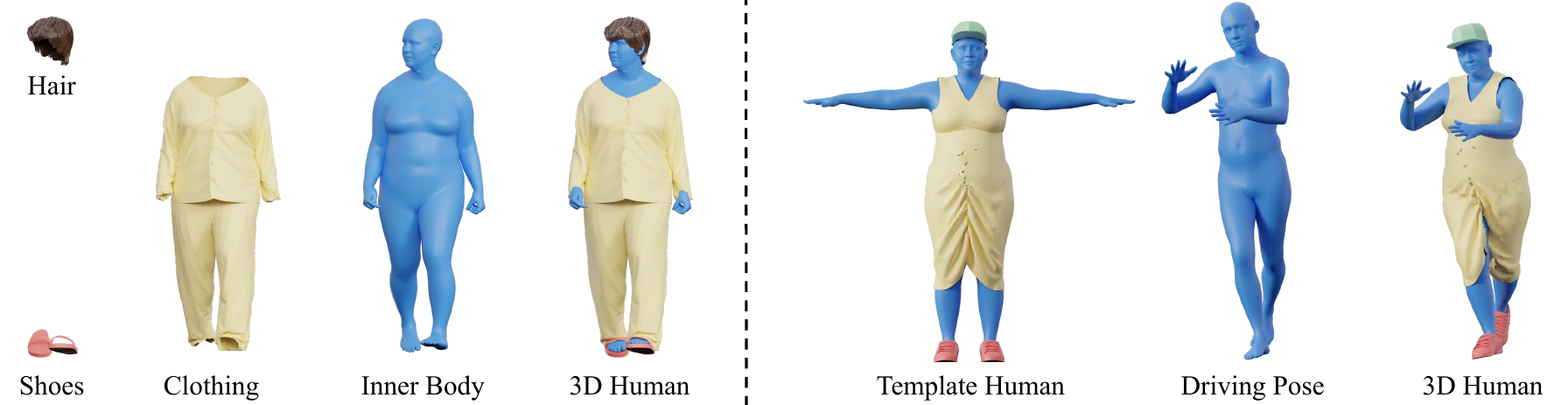}
    \caption{\scriptsize \textbf{Visual Examples from Our Two Synthetic Datasets.} \textbf{Left:} assets built on BEDLAM2~\cite{teschbedlam2}, supplemented with hair and shoe assets. \textbf{Right:} assets built on SynBody~\cite{yang2023synbody} driven by poses from Motion-X~\cite{lin2023motionx}.}
    \label{fig:appendix:syndata}
\end{figure*}

The second synthetic dataset is built based on SynBody~\cite{yang2023synbody} and Motion-X~\cite{lin2023motionx}. SynBody provides 1,000 3D human templates, each rigged with an inner body, clothing, and other accessories, which can be driven by \smplx pose parameters. We sample diverse poses from the Motion-X dataset and use them to animate the SynBody templates, resulting in 158,410 3D human assets with a wide variety of poses. Visual examples from both synthetic datasets are shown in~\cref{fig:appendix:syndata}.

\newpage

\section{Partial Data Reconstruction}

\begin{figure*}[h]
    \centering
    \includegraphics[trim=000mm 000mm 000mm 000mm, clip=true, width=\linewidth]{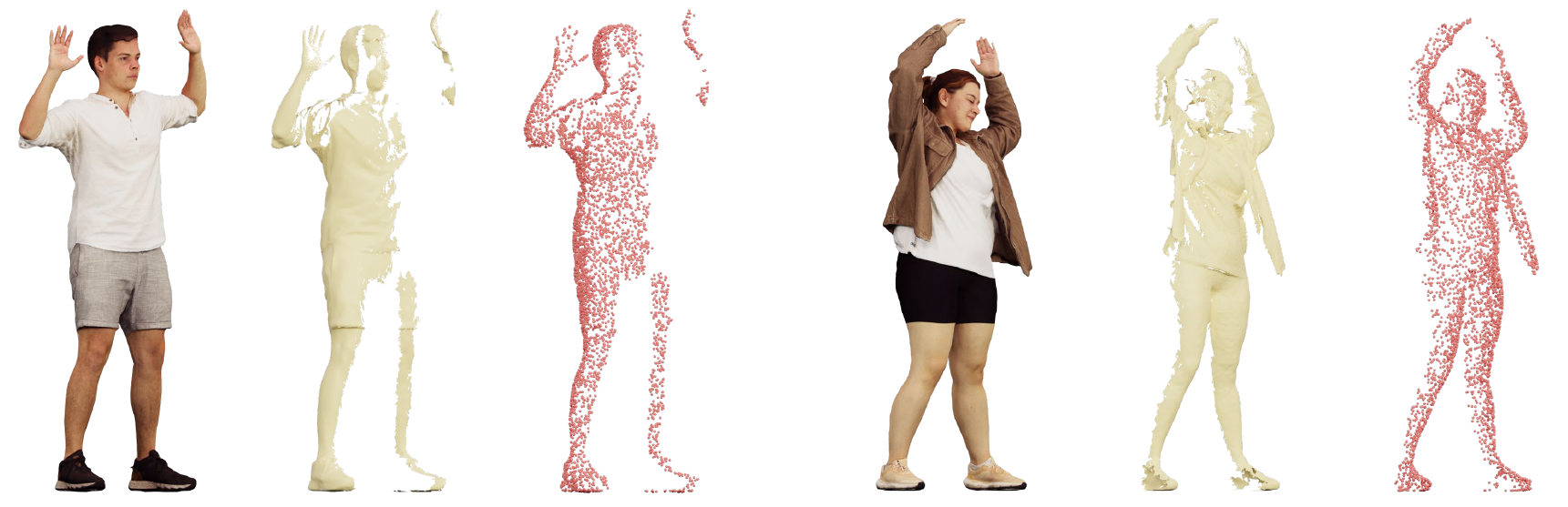}
    \caption{\scriptsize \textbf{Partial Data Reconstruction.} We simulate single-view point clouds by rendering a 3D asset from a random viewpoint, removing occluded faces and vertices, and sampling points on the remaining surface.}
    \label{fig:appendix:partial}
\end{figure*}

As described in Sec.~4.1 (Partial Data), we simulate single-view point clouds by rendering each 3D asset from a random viewpoint and sampling points on the visible surface. As illustrated in~\cref{fig:appendix:partial}, given a 3D human asset, we render it from a randomly sampled viewpoint and identify the indices of visible faces. The occluded faces and their corresponding vertices are then discarded, and points are sampled on the remaining mesh surface to produce the single-view point cloud. Incorporating such partial data enables \modelname to generalize to a broader range of input types, including human point clouds captured from depth sensors.

\newpage

\section{More Visual Results}

\begin{figure*}[h]
    \centering
    \includegraphics[trim=000mm 000mm 000mm 000mm, clip=true, width=\linewidth]{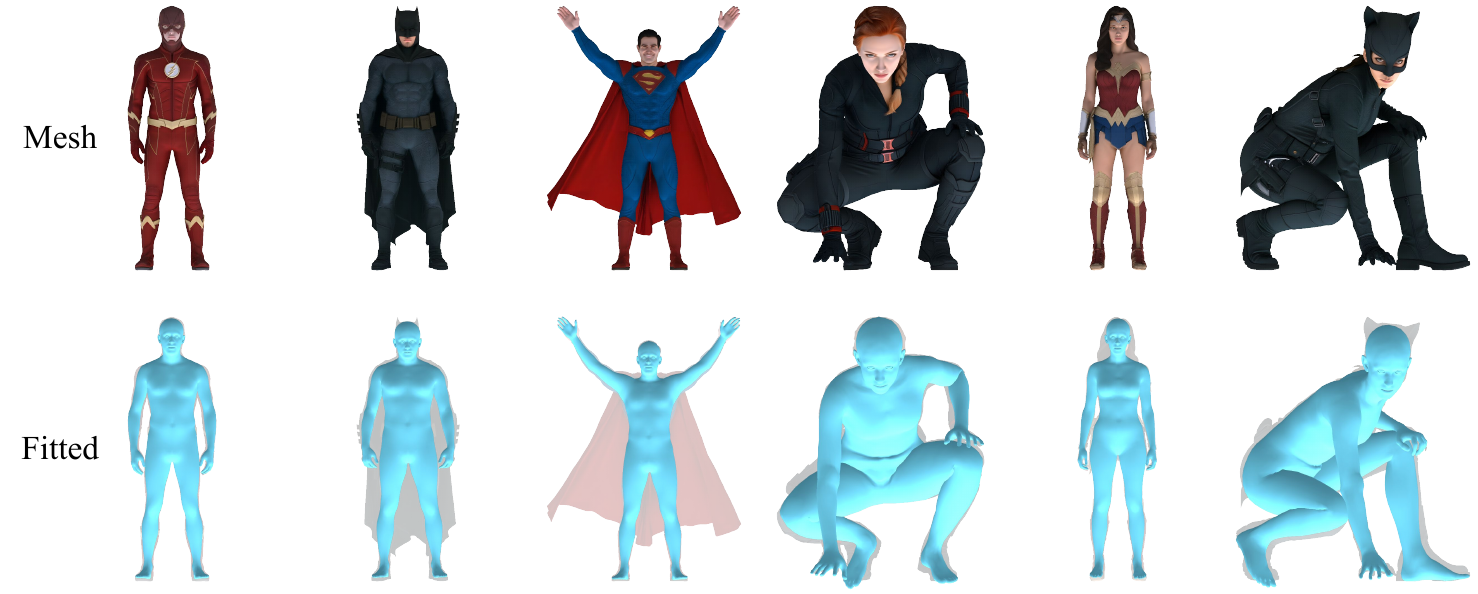}
    \caption{\scriptsize \textbf{Body Fitting Results on AI-generated Mesh.}}
    \vspace{-1.5em}
    \label{fig:appendix:aigc_res}
\end{figure*}

\pheading{Results on AI-generated 3D Human Assets.}
AI-generated 3D human assets often exhibit highly diverse clothing styles, poses, body shapes, and scale ambiguity, posing significant challenges for accurate 3D body fitting. Thanks to its strong generalization capability, \modelname effectively handles these challenges, as demonstrated in~\cref{fig:appendix:aigc_res}. Specifically, we obtain 3D human assets generated by Hunyuan3D~\cite{zhao2025hunyuan3d}, rescale them to the canonical human size using our scale predictor, and then feed them into the generalizable model and adapter for 3D body fitting.

\begin{figure}[p]
    \centering
    \includegraphics[trim=000mm 000mm 000mm 000mm, clip=true, width=\linewidth]{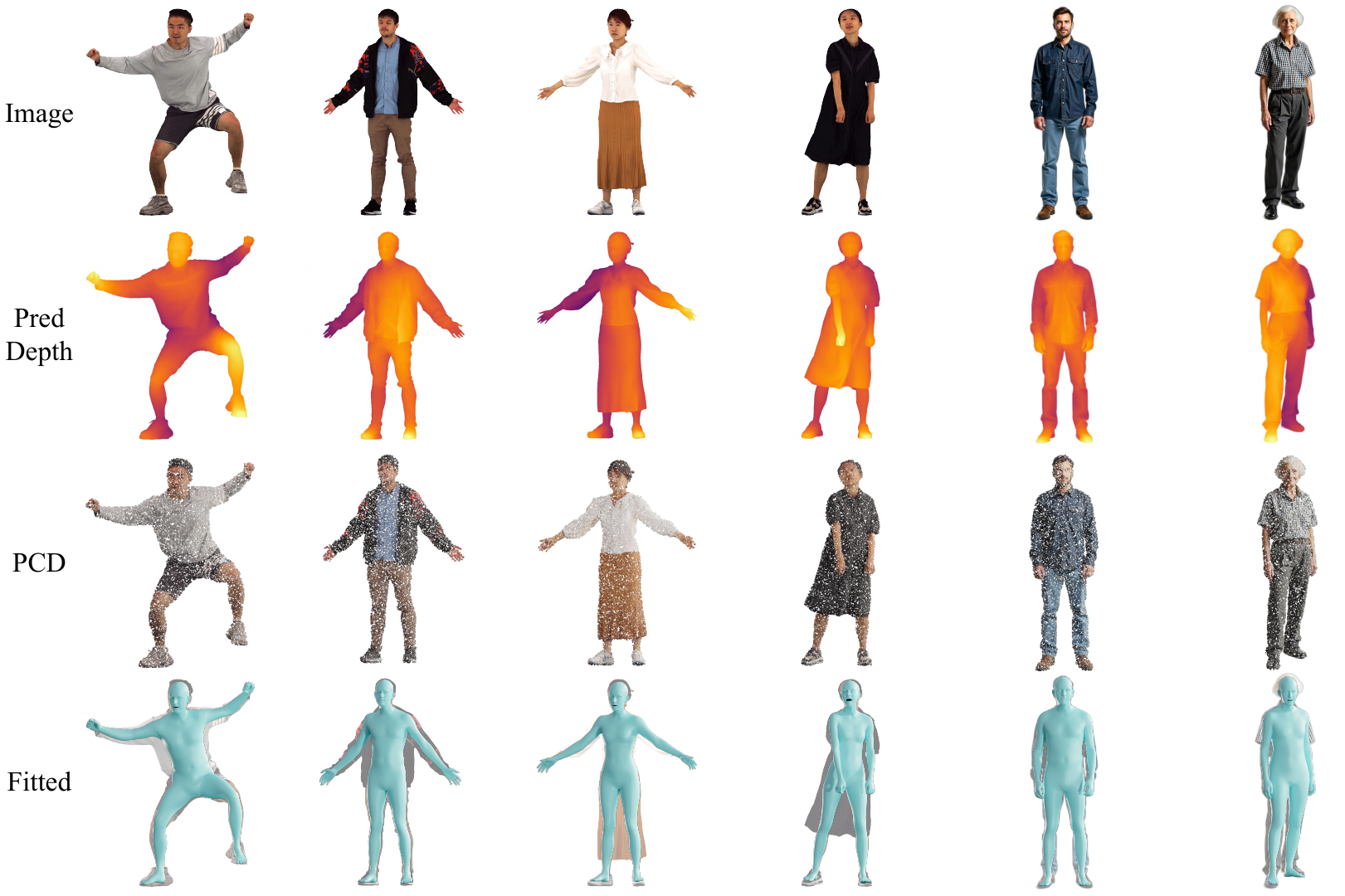}
    \caption{\scriptsize \textbf{Fitting Results on Depth Capture.} We estimate the depth map from a full-body image using Sapiens~\cite{khirodkar2024sapiens}, extract a point cloud, rescale and fit the 3D body with \modelname.}
    \label{fig:appendix:depth_capture}
    \vspace{1cm}
     \includegraphics[trim=000mm 000mm 000mm 000mm, clip=true, width=\linewidth]{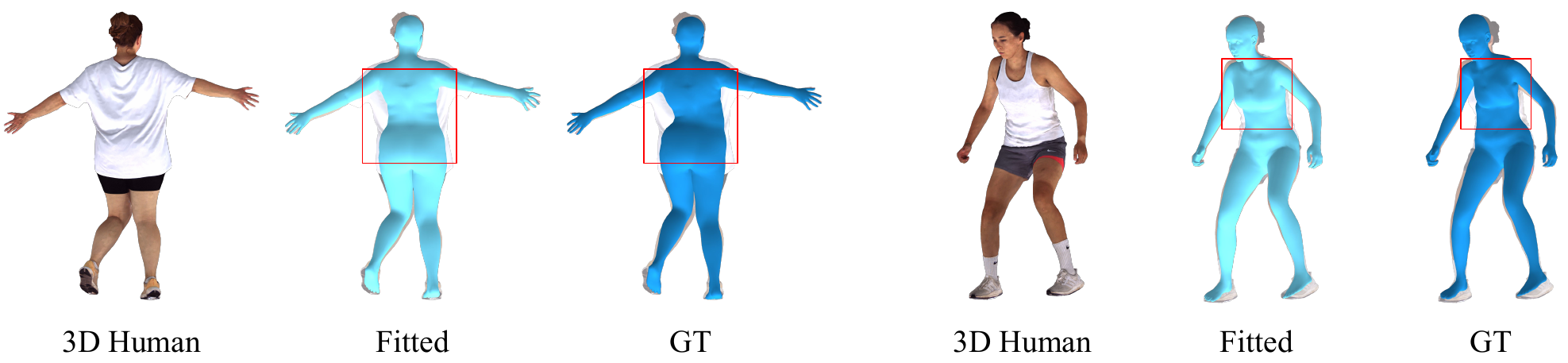}
    \caption{\scriptsize \textbf{Ours vs. Ground-truth.} Ground-truth bodies in \ddress are obtained by optimization, which can get stuck in local minima and produce unreasonable results. \modelname learns the human body distribution during training, leading to better estimates in som ceases.}
    \label{fig:appendix:better_than_gt}
\end{figure}

\pheading{Results on Depth Capture.}
\modelname can also be applied to point clouds captured from depth sensors. As shown in~\cref{fig:appendix:depth_capture}, given a full-body image, we use Sapiens~\cite{khirodkar2024sapiens} to estimate the depth map and extract a point cloud from it. The resulting point cloud is then passed through our scale predictor, generalizable landmark predictor, and adapter to perform 3D body fitting. The results confirm that our method generalizes well to this input modality, producing accurate body estimates from depth-captured data.

\pheading{Ours vs. Ground-truth.}
As a learning-based method, \modelname implicitly learns the distribution of human body shapes during training, yielding predictions that are more physically plausible and better conform to natural body structure. In contrast, the ground-truth annotations in \ddress are obtained via optimization-based fitting, which can converge to local minima and produce bad results in some cases. As shown in~\cref{fig:appendix:better_than_gt}, our method produces more reasonable 3D body estimates than the ground-truth in certain cases, highlighting the advantage of our learning-based approach.

\begin{figure*}[p]
    \centering
    \includegraphics[trim=000mm 000mm 000mm 000mm, clip=true, width=\linewidth]{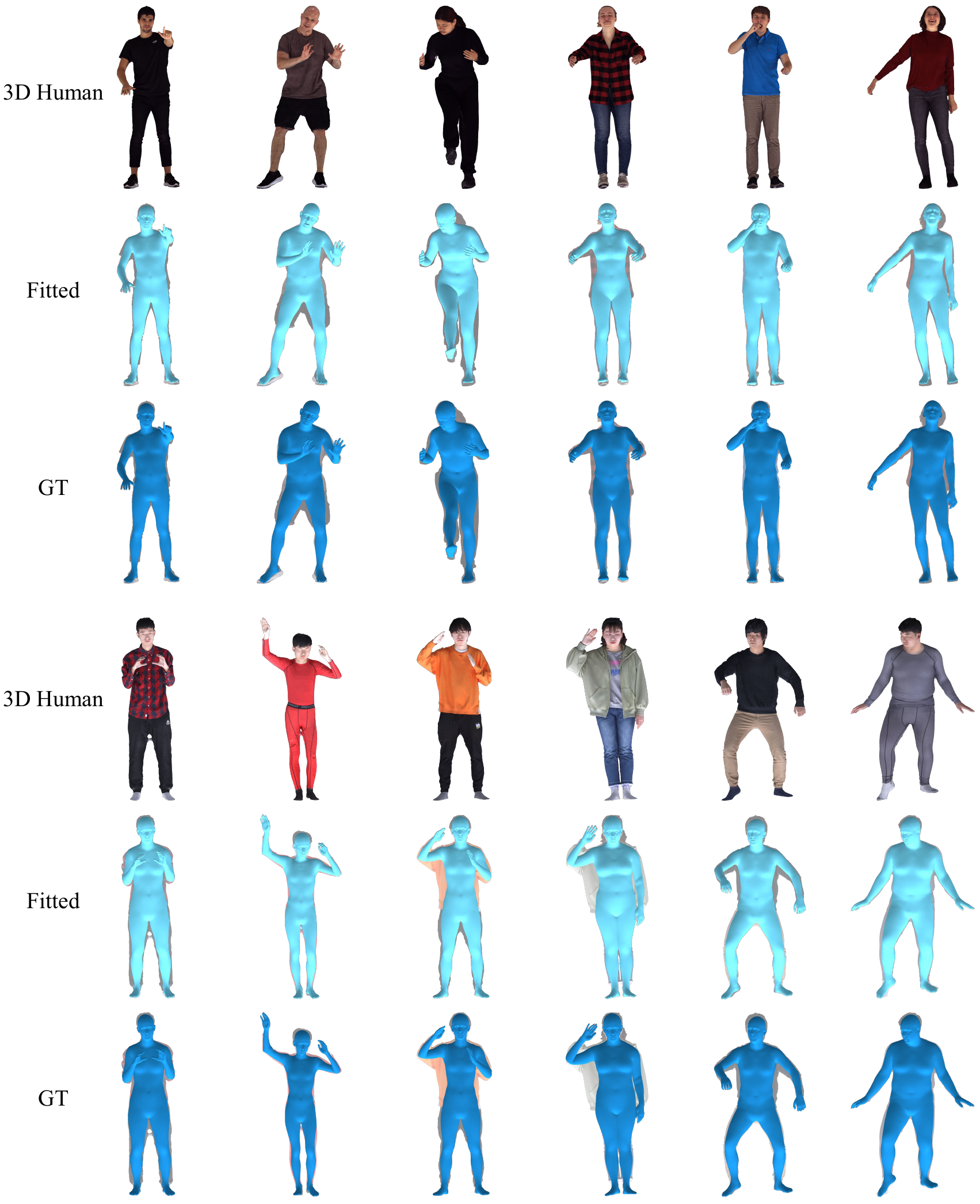}
    \caption{\scriptsize \textbf{More 3D Body Fitting Results.}}
    \label{fig:appendix:more_res}
\end{figure*}

\pheading{Additional 3D Body Fitting Results.}
We provide additional 3D body fitting results in~\cref{fig:appendix:more_res}, obtained using the generalizable model and adapter trained on the unified dataset. The results show that our method accurately recovers the underlying 3D body across a diverse range of clothing styles, poses, and shapes.


\section{Inference Time}
We evaluate the inference time of \modelname on a single NVIDIA RTX 4090 GPU (\cref{tab:appendix:infer_time}). With 5,000 input points, the scale predictor takes 0.066 seconds to estimate the scale factor. The point encoder takes 0.063 seconds to extract point features, and the landmark predictor takes 0.042 seconds to predict the 3D landmarks. If the image adapter is used, it takes an additional 0.007 seconds to extract image features and the landmark prediction time increases a slight to 0.021 seconds. Finally, the \smplx optimization step takes 9.088 seconds to fit the body model to the predicted landmarks.
\begin{table}[h]
    \centering
    \caption{\scriptsize \textbf{Inference Time (in Seconds) of Each Component in \modelname.}}
    \renewcommand{\arraystretch}{1.2}
    \label{tab:appendix:infer_time}
    \resizebox{0.8\textwidth}{!}{  
        \begin{tabular}{c|c|c|c|c}
            \bottomrule
            Scale Pred & Point Feat Pred & Image Feat Pred & Landmark Decode & SMPL-X Optim \\ 
            \hline
            \multirow{2}{*}{0.066} & \multirow{2}{*}{0.063} & $-$    & 0.017 & \multirow{2}{*}{9.088} \\
            \cline{3-4}
                                &                        & 0.007  & 0.021 &                        \\
            \toprule
        \end{tabular}
    }
\end{table}

\end{subappendices}

\end{document}